\pdfoutput=1
\documentclass[11pt]{article}

\usepackage{ACL2023}

\usepackage{times}
\usepackage{latexsym}
\usepackage{longtable}

\usepackage[T1]{fontenc}
\usepackage[utf8]{inputenc}

\usepackage{microtype}

\usepackage{inconsolata}

\usepackage{tabu}
\usepackage{graphicx} 
\usepackage{enumitem}
\usepackage{float} 
\usepackage{hyperref}
\usepackage{url}
\usepackage{booktabs}
\usepackage{multirow}
\usepackage{graphics}
\usepackage{amsmath}
\usepackage{multicol}
\usepackage{lipsum}
\usepackage{mwe}
\usepackage{hyperref}
\usepackage{url}
\usepackage{xcolor}
\usepackage[ruled,vlined]{algorithm2e}
\usepackage{bbm}
\usepackage{upgreek}
\usepackage{svg}
\usepackage{subfigure}
\usepackage{makecell}
\usepackage{dsfont}
\usepackage{fdsymbol}

\SetAlFnt{\small}
\usepackage{wrapfig}
\usepackage{colortbl}
\usepackage{bbm}
\usepackage{comment}

\newcommand{\brihi}[1]{{\color{blue} [{\bf Brihi}: #1]}}
\newcommand{\ziyi}[1]{{\color{brown} [{\bf Ziyi}: #1]}}
\newcommand{\zhewei}[1]{{\color{green} [{\bf Zhewei}: #1]}}
\newcommand{\aaron}[1]{{\color{cyan} [{\bf Aaron}: #1]}}
\newcommand{\ameya}[1]{{\color{green} [{\bf Ameya}: #1]}}
\newcommand{\sahana}[1]{{\color{brown} [{\bf Sahana}: #1]}}
\newcommand{\xiang}[1]{{\color{red} [{\bf Xiang}: #1]}}

\renewcommand{\brihi}[1]{}
\renewcommand{\ziyi}[1]{}
\renewcommand{\zhewei}[1]{}
\renewcommand{\aaron}[1]{}
\renewcommand{\ameya}[1]{}
\renewcommand{\sahana}[1]{}
\renewcommand{\xiang}[1]{}

\newcommand{\eg}{\textit{e.g., }}

\colorlet{lred}{red!15}
\colorlet{lblue}{blue!15}
\colorlet{lpink}{pink!75}
\colorlet{lyellow}{yellow!50}
\colorlet{lpurple}{purple!40}
\colorlet{lorange}{orange!40}
\colorlet{lgreen}{green!40}

\DeclareUnicodeCharacter{FFFD}{ }

\newcommand{\reward}{\textsc{Gen-U}\xspace}

\title{
Are Machine Rationales (Not) Useful to Humans? \\ Measuring and Improving Human Utility of Free-Text Rationales
}
\author{Brihi Joshi$^{\clubsuit}$\thanks{~~Equal contribution.} \hspace{3mm} Ziyi Liu$^{\clubsuit*}$ \hspace{3mm} Sahana Ramnath$^{\clubsuit}$ \hspace{3mm} \textbf{Aaron Chan}$^{\clubsuit}$ \hspace{3mm} \textbf{Zhewei Tong}$^{\diamondsuit}$ \hspace{3mm} \\ \textbf{Shaoliang Nie}$^{\spadesuit}$ \hspace{3mm}\textbf{Qifan Wang}$^{\spadesuit}$ \hspace{3mm} \textbf{Yejin Choi}$^{\vardiamondsuit\varheartsuit}$ \hspace{3mm} \textbf{Xiang Ren}$^{\clubsuit\varheartsuit}$ \vspace{1mm} \\
$^{\clubsuit}$University of Southern California \hspace{1mm} $^{\diamondsuit}$ Tsinghua University \hspace{1mm} $^{\spadesuit}$Meta AI \\ \hspace{1mm} $^{\varheartsuit}$Allen Institute for Artificial Intelligence \hspace{1mm} $^{\vardiamondsuit}$ University of Washington \\
\small{\texttt{\{brihijos, zliu2803, sramnath, chanaaro, xiangren\}@usc.edu}} , \small{\texttt{tzw19@mails.tsinghua.edu.cn}} \\
\small{\texttt{\{snie, wqfcr\}@meta.com}}, \small{\texttt{yejin@cs.washington.edu}}
}
\begin{document}
\maketitle
\begin{abstract}

% Recently, there has been a growing interest in using language models (LMs) for human-AI collaboration.
% To explain their reasoning processes to humans, state-of-the-art LMs have been shown to generate free-text rationales fluently, \eg via chain-of-thought prompting. 
% % Still, it remains unclear how effectively these generated FTRs can provide \textit{utility} for human-AI collaboration, \ie assist lay humans in solving and learning challenging NLP tasks.  

Among the remarkable emergent capabilities of large language models (LMs) is free-text rationalization; beyond a certain scale, large LMs are capable of generating seemingly useful rationalizations, which in turn, can dramatically enhance their performances on leaderboards. This phenomenon raises a question: can machine generated rationales also be useful for humans, especially when lay humans try to answer questions based on those machine rationales?
% In this paper, we investigate how effectively these generated rationales can provide \textit{utility} in a human-AI collaboration setting, \ie assisting lay humans in solving and learning a challenging task.  
% \sahana{We formally define \textit{human-utility} as . . .}.
We observe that human utility of existing rationales is far from satisfactory, and expensive to estimate with human studies.
Existing metrics like task performance of the LM generating the rationales, or similarity between generated and gold rationales are not good indicators of their human utility.
While we observe that certain properties of rationales like conciseness and novelty are correlated with their human utility, estimating them without human involvement is challenging.
We show that, by estimating a rationale's helpfulness in \textit{answering similar unseen instances}, we can measure its human utility to a better extent.
We also translate this finding into an automated score, \reward, that we propose, which can help improve LMs' ability to generate rationales with better human utility, while maintaining most of its task performance.
Lastly, we release all code and collected data with this project.\footnote{\url{https://github.com/INK-USC/RationaleHumanUtility}}
% \xiang{should add a 2nd sentence in abs to speak out “why we care human utility” besides other use of the FTRs.}
% In this work, we ask: can these rationales also contribute to \textit{improving human utility} of accomplishing a given task?
% To investigate what makes an FTR useful to humans, this paper analyzes the relationships between human utility and various LM/FTR properties.
% First, although LMs are often finetuned/prompted to generate task labels and FTRs jointly, we find that LMs' task performance has little correlation with human utility of the FTRs, whereas LM size is a positive predictor of human utility.
% Second, we observe that certain FTR property pairs are strong positive predictors of human utility, \eg high-utility FTRs tend to both be concise and contain novel information.
% Third, we show that high-utility FTRs for a given task instance can provide transferable knowledge that helps humans generalize to solving new instances.
% We also observe that a high utility rationale for a given instance contains enough context to help humans generalize to even unseen instances, for which they do not have access to any rationales.
% By shedding light on the nature of FTRs' human utility in practical settings, our findings can help guide future work on designing LMs and FTR generation strategies for stronger human-AI collaboration.
% We believe that investigating human utility of rationales can better situate LMs in real-world scenarios, where human-LM collaboration is required to improve both human understanding of the task as well as LM performance.

\end{abstract}

% \DeclareUnicodeCharacter{U+FFFD}{}

% \renewcommand{\brihi}[1]{}
% \renewcommand{\ziyi}[1]{}
% \renewcommand{\zhewei}[1]{}
% \renewcommand{\aaron}[1]{}
% \renewcommand{\ameya}[1]{}
% \renewcommand{\sahana}[1]{}
% \renewcommand{\xiang}[1]{}

\section{Introduction}
% \brihi{Need to restructure better}
% Points to mention in the introduction - 
% 1. Talk about LMs, their transparency, extractive and FTRs
% 2. Improvement in task performance
% 3. Constraints in improvements -- what causes improvements, rationales not necessarily good
% 4. Finding 1 --> across model, distinct model
% 5. Finding 2 --> Property correlations

In recent years, there has been a surge of interest in using language models (LMs) for human-AI collaboration \cite{wiegreffe-etal-2022-reframing, you-lowd-2022-towards}.
\begin{figure}[h!]
% \vspace{-0.9cm}
    \centering
    \includegraphics[width=0.95\columnwidth]{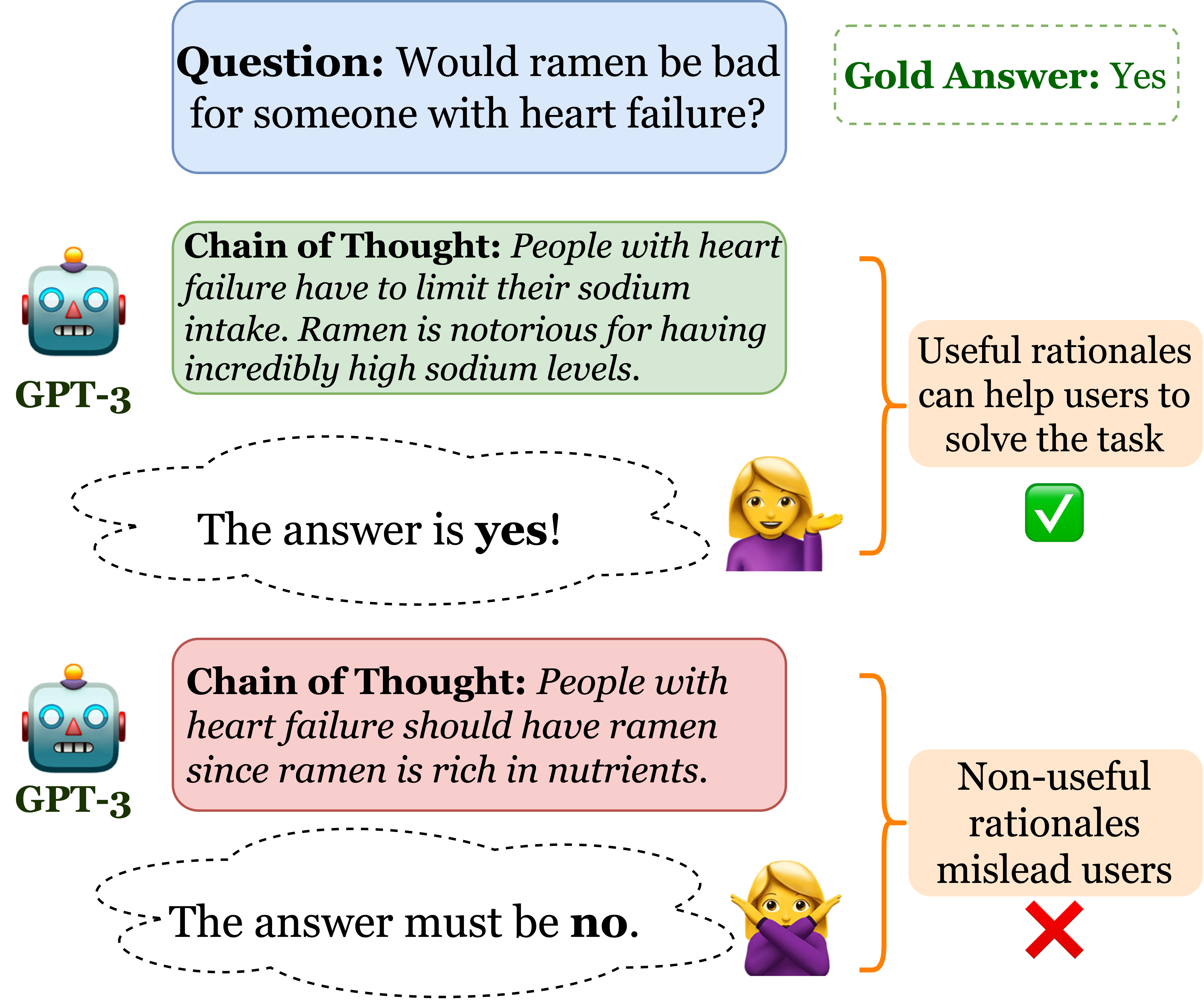}
    % \caption{\textbf{An illustration of our pipeline to evaluate human utility of free-text rationales:} Humans answer a question without a rationale. An LM then shows them a rationale, that helps them correctly answer the same question, after which they rate the rationale across different axes like \textit{novelty} (shown here). The same rationale also helps them answer a new question which uses a similar reasoning process as the original question.}
    % \vspace{-0.2cm}
    \caption{\textbf{An illustration of Human Utility of rationales:} Here, we show Chains of Thought (rationales) generated by GPT-3 in two scenarios. The first one is providing knowledge to the human to be able to answer the question, but the second rationale is not useful, and is in fact, misleading the human to answer incorrectly.}
    \label{fig:utility_intro}
\vspace{-0.3cm}
\end{figure}
\begin{figure*}[h!]
% \vspace{-0.8cm}
    \centering
    \includegraphics[width=0.98\linewidth]{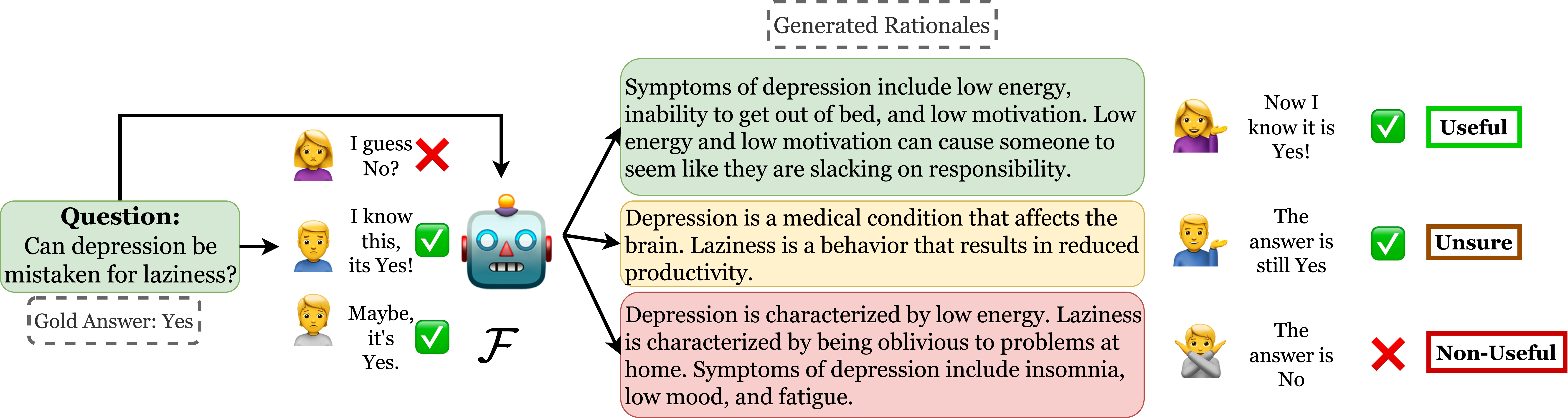}
% \vspace{-0.6cm}
    \caption{\textbf{An illustration of measuring human utility of machine rationales.} We evaluate whether a human's belief of the answer changes before and after seeing a rationale generated by an LM.}
    \label{fig:utility_pipeline}
\vspace{-0.3cm}
\end{figure*}
For example, LMs have played a large role in reducing human effort for dataset creation \cite{bonifacio2022inpars, yuan2021synthbio, wanli} and helping humans critique text \cite{saunders2022self}. 
However, the opaque reasoning processes of these LMs pose serious concerns about their role in high-stakes decision-making \citep{bender2021dangers, doshi2017towards}.
Recently, many works have explored using LMs to generate fluent, human-like \textit{free-text rationales}\footnote{We use the term `\textit{rationales}' throughout the paper to refer to free-text rationales and explanations.} via natural language \cite{ehsan2018rationalisation, rajani2019explain} that can explain their decisions.
% While extractive explanations highlight important input tokens \cite{sundararajan2017axiomatic, li2016understanding, chan2022unirex}, rationales can fluently provide human-like explanations via natural language \cite{ehsan2018rationalisation, narang2020wt5, rajani2019explain, esnli, cot, majumder2021knowledge}.
Further, rationales can reference things beyond the task input, and also support high flexibility in content, style, and length \cite{narang2020wt5, wiegreffe-etal-2022-reframing, wiegreffe-etal-2021-measuring, chan2022frame}. 
However, evaluating if a rationale of a task-instance contains enough knowledge to help lay humans understand and solve that instance correctly is still under-explored.
Prior literature for human-AI collaboration has studied plausibility \cite{dataset_review}. 
% which measures how well the generated rationales align with human-provided gold rationales. 
However, plausibility only aims to capture human judgement of the rationale supporting LM's predicted label.
There has been little work done on evaluating actionable advantages offered by rationales to \textit{lay humans} in understanding a task, despite the promise of human-AI collaboration \cite{schuff2022how}.
Studying human utility of rationales is important to not only situate them in real-world use cases beyond the involvement of researchers, but also to bridge the gap between human and AI understanding, specifically in scenarios where AI systems perform better.
In this work, we shift the paradigm of rationale evaluation, by investigating \textit{human utility} of rationales in helping lay humans understand and solve a given task correctly.

In our study, we observe that \textit{human utility of current LMs is far from satisfactory (including large LMs like GPT-3)}, with only $20\%$ of generated rationales being actually useful (\textsection \ref{sec:utility_setup}).
% First, \textit{do existing approaches generate rationales with high human utility?}( \brihi{Add section ref}) 
% LMs are often finetuned/prompted to jointly generate task labels and rationales.
% However, , 
Given that human evaluations are expensive, we should find a reliable way to measure human utility.
% Nonetheless, across a wide range of LM architectures and prompt templates, our human-subject studies show that a very few percentage of generate rationales are actually beneficial for humans in solving the task correctly.
% Furthermore, evaluating human utility requires human-subject evaluations, which are often tedious and expensive.
We examine the correlation of two straightforward measures like LM task performance and alignment with gold rationales, with human utility and find no usable insights.
% We observe that there is no correlation between an FTR's human utility and metrics like the task performance of the LM that generated it, or similarity between generated and gold FTR.
We also ask humans to evaluate rationales w.r.t eight granular-level properties.
While we observe that six out of these eight properties are correlated with human utility, reliably estimating them without human evaluation is still an open question \cite{roscoe}. 
% We also ask humans to evaluate both LM-generated and gold FTRs with respect to eight FTR properties: grammaticality, validity, coherence, conciseness, leakage, novelty, association, and contrast.

% First, \textit{what is the correlation between an FTR's human utility and its corresponding LM's task performance?} (\textsection \ref{sec:perf_util})
% LMs are often finetuned/prompted to jointly generate task labels and FTRs.
% Nonetheless, across a wide range of LM architectures and prompt templates, our human-subject studies show little correlation between an FTR's human utility and the task performance of the LM that generated it.
% Furthermore, we find that larger LMs (\eg GPT-3) tend to generate FTRs with higher human utility, while LM finetuning is a more important factor in determining the LM's task performance.

% We investigate whether task performance of LMs that generate FTRs is correlated with human utility of these FTRs.
% Through human-subject studies, we observe no correlation \sahana{soften/elaborate this statement? it seems pretty harsh for the intro}.
% Furthermore, we observe that larger LMs (like GPT-3) often generate more useful FTRs, while not requiring fine-tuning with gold FTRs, as compared to smaller LMs (T5) which often have a higher task performance (due to fine-tuning). \sahana{The second half of this sentence is straying away from the point.}

In addition to the above observation, we find that \textit{high-utility rationales effectively transfer knowledge to humans for solving new instances.} (\textsection \ref{sec:generalization})
% Second, \textit{to what extent do FTRs for a given task instance help humans generalize to new instances?} (\brihi{Add section ref})
We create new instances (\eg questions) by either paraphrasing the original instance in a nontrivial manner (rephrase), editing the original instance so that its correct label is changed (counterfactual), or writing an instance that requires a similar reasoning process as the original instance (similar reasoning).
We observe that useful rationales help humans generalize better to new instances, whereas non-helpful rationales even mislead them to answer incorrectly.
To follow up on the above finding, we show that we can \textit{improve an LM's ability to generate rationales with better human utility.} (\textsection \ref{sec:genlm})
We translate this finding into an automated score, \reward, that reflects the ability of a rationale to help an LM answer generalization instances, that better correlates with human utility (when compared to other metrics like LMs' task accuracy).
% Third, \textit{can we improve LMs' ability to generate rationales with better human utility?} (\brihi{Add section ref})
We use \reward as a reward \cite{lu2022quark} while generating rationales and observe that the updated LM generates $2\%$ more useful rationales and gets rid of $4\%$ misleading rationales than before, via human-subject evaluations, without hindering the LMs' task performance.

This paper presents the first comprehensive study of lay human utility of free-text rationales.
By introducing lay humans in the rationale evaluation pipeline, and using their insights into existing LMs, we believe our findings can help guide future work on developing methods for efficient and reliable human-AI collaboration.
% This paper presents the first comprehensive study of why humans perceive some FTRs to be more useful than others.
% By analyzing the relationships between FTRs' human utility and various LM/FTR properties, we establish a better understanding of how LMs and FTR generation strategies can be designed to yield higher human utility.
% We believe our findings can help guide future work on developing methods for efficient and reliable human-AI collaboration.
% why useful TBD

\section{Human Utility of LM-generated Rationales}
\label{sec:utility_setup}

\begin{table*}[h]
\vspace{-0.8cm}
\centering
\scalebox{0.63}{
\begin{tabular}{ll}
\toprule
\textbf{Human Utility} & \textbf{Examples} \\
\midrule
\textbf{\textsc{Useful}}&
\makecell[l]{
\textbf{\textsc{Question}}: Could Al Capone have read works from the Harlem Renaissance? \textbf{\textsc{Gold Answer}}: Yes\\
\textbf{\textsc{Rationale}}: The Harlem Renaissance began in the late 1600's. Al Capone was born in 1888.\\
\midrule
\textbf{\textsc{Question}}: Has Alan Greenspan lived through at least a baker's dozen of president's? \textbf{\textsc{Gold Answer}}: Yes\\
\textbf{\textsc{Rationale}}: Alan Greenspan was born in 1926. He has lived through the presidencies of Roosevelt, Truman, Eisenhower, Kennedy, Johnson,\\ Nixon, Ford, Carter, Reagan, Bush, Clinton, Bush, Obama, Trump.
\\
% \textbf{Generalization Question}:Is Trent Reznor associated with David Lynch?
}\\
\midrule
\textbf{\textsc{Unsure}}&
\makecell[l]{
\textbf{\textsc{Question}}: Can a sea turtle play tennis using a tennis racket? \textbf{\textsc{Gold Answer}}: No \\
\textbf{\textsc{Rationale}}: Sea turtles are aquatic animals that cannot swim well under water. Tennis rackets are designed for land animals such as humans.\\
\midrule
\textbf{\textsc{Question}}: Do gorillas fight with panda bears? 
\textbf{\textsc{Gold Answer}}: No \\
\textbf{\textsc{Rationale}}: Gorillas are native to Africa. Panda bears are native to China. \\
}\\

% \textbf{Useful}&
% \makecell[l]{
% \textbf{Question}: Could Al Capone have read works from the Harlem Renaissance?\\
% \textbf{Answer}: Yes \\
% \textbf{Rationale}: The Harlem Renaissance began in the late 1600's. Al Capone was born in 1888.\\

% \midrule
% \textbf{Question}: Has Alan Greenspan lived through at least a baker's dozen of president's?\\
% \textbf{Answer}: Yes \\
% \textbf{Rationale}: Alan Greenspan was born in 1926. He has lived through the presidencies of Roosevelt, Truman, Eisenhower, Kennedy, Johnson,\\ Nixon, Ford, Carter, Reagan, Bush, Clinton, Bush, Obama, Trump.

% }
% \\
% \midrule
% \textbf{Unsure}&
% \makecell[l]{
% \textbf{Question}: Can a sea turtle play tennis using a tennis racket?\\
% \textbf{Answer}:No \\
% \textbf{Rationale}: Sea turtles are aquatic animals that cannot swim well under water. Tennis rackets are designed for land animals such as humans.\\
% \midrule
% \textbf{Question}:  Do gorillas fight with panda bears?\\
% \textbf{Answer}:No\\
% \textbf{Rationale}: Gorillas are native to Africa. Panda bears are native to China.
% }\\
\midrule
\textbf{\textsc{Not Useful}}&
\makecell[l]{
\textbf{\textsc{Question}}: Does 50 Cent get along with Jeffrey Atkins? 
\textbf{\textsc{Gold Answer}}: No\\
\textbf{\textsc{Rationale}}: Jeff Atkins is a musician who has recorded with 50 Cent. Jeff Atkins has a reputation for being a hard worker and a good friend.\\
\midrule
\textbf{\textsc{Question}}:  Do Apollo and Baldur share similar interests? 
\textbf{\textsc{Gold Answer}}: Yes\\
\textbf{\textsc{Rationale}}: Apollo is the god of the sun and the arts. Baldur is the god of light and peace.\\
}\\

\bottomrule
\end{tabular}
}
\caption{\textbf{Examples of rationales with different human utility from the StrategyQA Dataset:} Shown here are questions, rationales and gold answers, for different rationale types, as evaluated by our human studies.}
\label{tab:examples_utility}
\vspace{-0.3cm}
\end{table*}

We begin by defining human utility, intuitively and formally, and describing the LMs that we use for the rest of the paper.
Based on this definition, we conduct human studies to investigate whether existing LMs are capable of generating useful rationales.
Finally, we follow this up by identifying granular-level syntactic and semantic properties of rationales can indicate their human utility.
% \xiang{(1) A lot is going on in this section. Good to have an overview paragraph to lay out the storyline and establish the transition/relationship between different parts.
% (2) Add transition sentences at the beginning of each part.}
% In this section, we investigate the human utility of free-text rationales in solving tasks. We first defined human utility and its calculation methods; then we discussed the models we used to generate free-text rationales; at last we evaluated the human utility and properties of rationales.
% \paragraph{What is Human Utility?}
\paragraph{Human Utility of Machine Rationales.}
% \brihi{Update paragraph with new defition of utility considering I-O}
% We define human utility as the \textit{ability of a human to be able to solve the task correctly} \cite{idahl2021towards}. 
% It is shown that explainability methods are often used to debug models \cite{bhatt2019explainable} by practitioners, rather than facilitating trust between a model and a human.
% Note the difference between our definition of utility with that of forward simulation \cite{rigorous}. 
% Forward simulation requires a human prediction to match with that of the model, regardless of the true answer, and is more correlated with establishing trust between the model and the human.
We first define human utility of rationales as \textit{the advantage that rationales offer lay humans to solve tasks, that they are otherwise unable to} \cite{schuff2022how, idahl-etal-2021-towards, chu2020arevisual} (Figure \ref{fig:utility_pipeline}). 
In theory, we can estimate human utility of a rationale in a forward simulation-like \cite{doshi2017towards} setup: the difference in human performance of a task, with and without the assistance of a rationale.
% \textit{the ability of a human to correctly solve a task with a rationale, that they are otherwise unable to} \cite{idahl-etal-2021-towards, chu2020arevisual}. 
% We can measure the human utility of a rationale \sahana{on the basis of} \textit{if the rationale helps the human solve the task correctly, when the human solves the task incorrectly without the rationale}.
In this work, we reformulate this definition of utility for a classification task (multi-choice question answering). 
We use the StrategyQA \cite{geva-etal-2021-aristotle} and OBQA \cite{Mihaylov2018CanAS} datasets for our paper. 
The reason for doing so is to pick tasks where humans are not already better than LMs (unlike NLI and CommonsenseQA \cite{nangia2019muppet, talmor2021commonsenseqa}), and study cases where rationales are capable of knowledge transfer that can help humans.
More details about our task and dataset selection reasoning is highlighted in \textsection \ref{sec:apx:task_dataset_selection}.

% \brihi{Write short decription of the task, and link to appendix}.
\paragraph{Formal setup for calculating human utility.} 

Let $\mathcal{F}$ be a \textit{self-rationalizing LM} \cite{wiegreffe2020measuring} that can generate rationales with its predictions, and a corresponding input-output pair ${x, y}$.
$\mathcal{F}$ takes in $x$ as an input and generates a prediction $y_p$, and a rationale that corresponds to this prediction $r_p$.
% The task accuracy (which corresponds to accuracy, when aggregated over all the instances) of this instance is given as follows: 
% $$ \small \textsc{Task Accuracy} = \begin{cases} 
%       1 & y_p = y \\
%       0 & otherwise
%    \end{cases}
% $$

Let $\mathcal{H}$ be a human predictor that first takes in the instance $x$ and predicts a label for that instance, $y_h$.
Then, $\mathcal{H}$ is also shown the rationale $r_p$ and now takes both the instance and rationale ${x, r_p}$ as an input, and predicts a label $y_{hr}$.
Therefore, human utility of the rationale $r_p$ is calculated as:
$$ \small \textsc{Human Utility} = \begin{cases} 
      \textsc{Useful} & y_h \neq y\ \& \ y_{hr} = y\\
      \textsc{Not Useful} & y_{hr} \neq y\\
      \textsc{Unsure} & y_h = y\ \& \ y_{hr} = y
   \end{cases}
$$

In other words, rationales are \textit{useful} if a human incorrectly solved the task before, and with the introduction of the rationale, is able to correct their answer.
If even after being shown the rationale, the human is still solving the task incorrectly, this implies that the rationale has \textit{not} been useful.
However, if the human was correct both before and after being shown the rationale, we cannot conclusively determine the role of the rationale in helping solve the task.
We term these rationales as \textit{unsure}.
This category of instances can either be too easy, or it can be the case that the human was already aware of the answer even before being shown the rationale. 
% \brihi{See if examples of unsure/useful/not useful rationales can be added to appendix}.
Of course, this can also imply that the rationale has still been useful in answering the task correctly, however, our definition of utility specifically evaluates cases where rationales are solely responsible for human utility.

\begin{figure*}[h!]
\vspace{-0.9cm}
    \centering
    \includegraphics[width=0.93\linewidth]{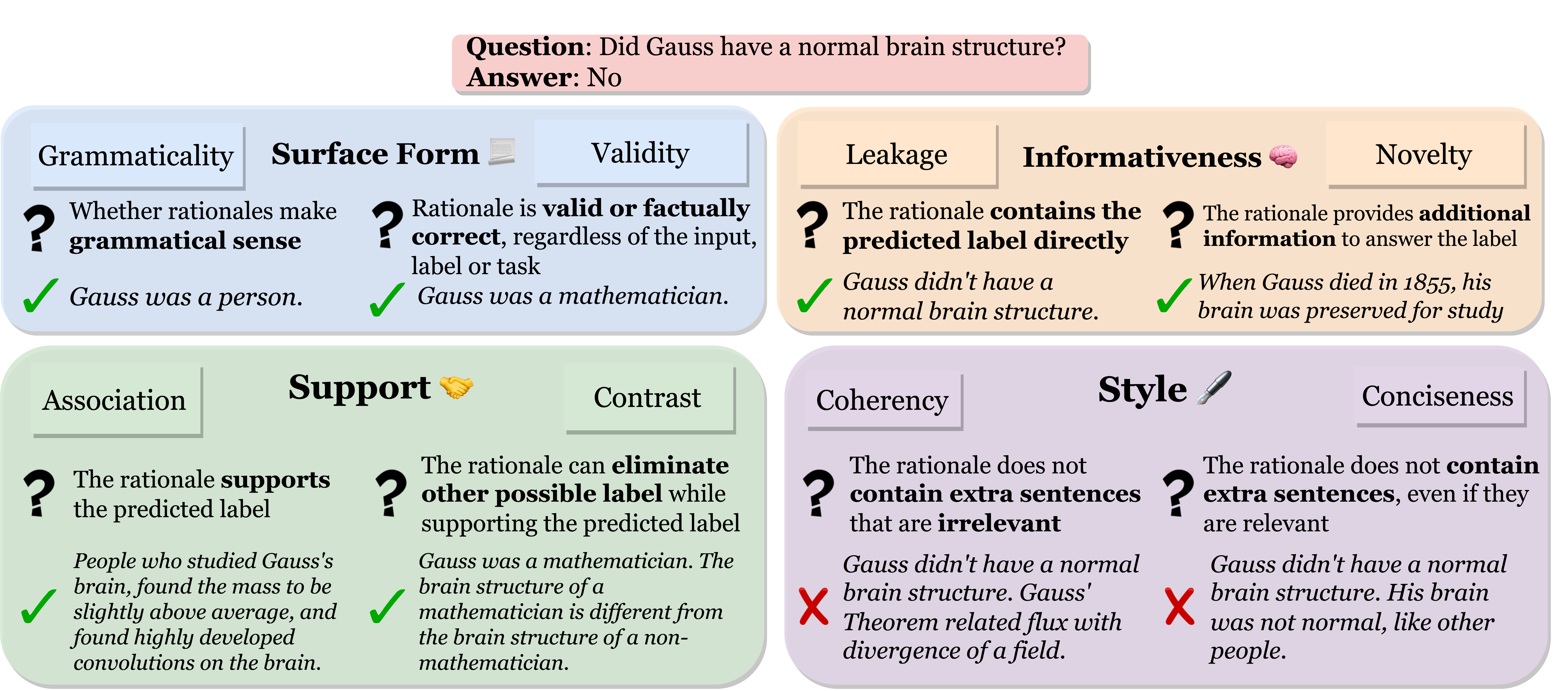}
    \caption{\textbf{Granular-level Rationale Properties:} Definitions for properties along each axes (surface form, informativeness, support and style) are shown. For all but style axes, an example of a rationale \textit{satisfying} the property is also shown. For style, we show examples of rationales that \textit{do not satisfy} the given properties.}
    \label{fig:properties}
    \vspace{-0.3cm}
\end{figure*}

% \subsection{Defining Utility Properties}
% \paragraph{Task and Dataset Selection.}
\paragraph{Self-rationalizing Models.} 
For our choice of $\mathcal{F}$, we experiment with in-context learning and fine-tuning based approaches. 
For the rest of our paper, we pick three LM configurations that provide us the best task accuracy for the rest of our experiments in this paper: davinci-instruct-beta (GPT3) \cite{brown2020language} with six randomly picked demonstrations, with the FEB \cite{marasovic-etal-2022-shot} template, where rationales are generated after the predicted answer, T5-large with full fine-tuning and infilling template \cite{marasovic-etal-2022-shot} and T5-3B with 128-shot fine-tuning and infilling template.
Details about prompt templates, experiment settings and model selection are in \textsection \ref{sec:apx:self_rationalising_models}.
% \xiang{Nice to have a small table summarizing the couple best performing models' perf.}
\begin{table}[h!]
\centering
\scalebox{0.70}{
\begin{tabular}{ccccc}
\toprule

Dataset & Model & Setting & Test accuracy\\
 
\midrule
\multirow{3}{*}{\textsc{StrategyQA}} & \textsc{T5-large}&full-finetuning&67.03\\
&\textsc{T5-3B}&128-shot&56.70$\pm$1.85\\
&\textsc{GPT-3-175B}&in-context&60.04\\
\midrule
\multirow{3}{*}{\textsc{OBQA}} & \textsc{T5-large}&full-finetuning&65.72\\
&\textsc{T5-3B}&128-shot&56.70$\pm$1.85\\
&\textsc{GPT-3-175B} &in-context&55.60\\
\bottomrule
\end{tabular}
}
\caption{\textbf{Self-Rationalising Model Results}: Shown here are the test set accuracies of T5-Large, T3-3B and davinci-instruct-beta (GPT-3) from best settings. We use these three settings for the rest of our work. The results of the complete list of finetuning and in-context learning experiments we performed are shown in Tables \ref{tab:finetune_res}, \ref{tab:icl_res} and \ref{tab:icl_res_obqa}.}
\label{tab:best_perf_model}
\vspace{-0.5cm}
\end{table}
\paragraph{To what extent do LM-generated rationales provide utility to humans?}

% \xiang{mention/quote numbers on pairwise/inter-annotator agreement.}
We conduct human-subject studies to evaluate utility of free-text rationales. 
We use Amazon Mechanical Turk \footnote{\href{www.mturk.com}{www.mturk.com}} to first curate a set of annotators that understand the task well (via extensive qualification tests). 
Each instance is answered by five annotators. (The annotator agreements are shown in Table \ref{tab:human_agreement}).
For each StrategyQA and OBQA test instance, we ask humans to first provide an answer given the question.
We then show them a rationale and ask them to answer the question again.
The rationale shown to them is generated by either of the three selected LMs.
Details about MTurk experiment setup and annotation agreements are in \textsection \ref{sec:apx:mturk}.
For each instance, we calculate human utility as defined above, where predictions made by five annotators are aggregated by taking a majority vote. 
% \ziyi{The annotation agreement is shown in Table }\ref{tab:human_agreement}.

\begin{table}[h!]
\scalebox{0.63}{
\begin{tabular}{clcccc}
\toprule
& &\multicolumn{4}{c}{\% of generated rationales} \\
\cmidrule{3-6}
Dataset & Type & All & GPT-3-175B   & T5-3B   & T5-Large \\
\midrule
\multirow{3}{*}{StrategyQA} &\textsc{Useful} & 17.83 & 20.30 & 18.12 & 15.06  \\
&\textsc{Not Useful} & 35.00 & 25.76 & 35.15 & 44.10  \\
&\textsc{Unsure}     & 47.16 & 53.93 & 46.72 & 40.82  \\
\midrule
\multirow{3}{*}{OBQA} &\textsc{Useful}& 15.26 & 16.06 & 14.85 & 14.85  \\
&\textsc{Not Useful} & 54.88 & 54.21 & 50.60 & 59.83  \\
&\textsc{Unsure}     & 29.85 & 29.71 & 34.53 & 25.30  \\
\bottomrule
\end{tabular}
}
\caption{\textbf{Distribution of Human Utility of Rationales:} Shown here are the \%s of different types of rationales based on their utility, for T5-Large, T5-3B and davinci-instruct-beta (GPT-3), for both StrategyQA and OBQA.}
\label{tab:percent_useful}
\vspace{-0.1cm}
\end{table}

We observe that (Table \ref{tab:percent_useful}) for all the LMs combined only a small amount of rationales generated are actually useful for humans.
A large chunk of rationales also mislead humans to select the incorrect answer (\textsc{Not Useful}).
In fact, for T5-Large and UnifiedQA-Large, the configuration that led to the best task performance for StrategyQA and OBQA, has the highest \% of \textsc{Not Useful} rationales.

\paragraph{Do existing metrics correlate with human utility?}

\begin{table}[h!]
\scalebox{0.59}{
\begin{tabular}{clcccc}
\toprule
& &\multicolumn{4}{c}{Correlation} \\
\cmidrule{3-6}
Dataset & Type & Overall & GPT-3-175B & T5-3B & T5-Large \\
\midrule
\multirow{2}{*}{\textsc{StrategyQA}} & \textsc{Task Accuracy} & 0.035 & \textbf{0.111} & 0.034 & 0.005  \\
& \textsc{BERTScore} & 0.041 & \textbf{0.021} & 0.017 & 0.002  \\
% &\textsc{Unsure}     & 47.16 & 53.93 & 46.72 & 40.82  \\
\midrule
\multirow{2}{*}{\textsc{OBQA}} & \textsc{Task Accuracy} & 0.022 & \textbf{0.092} & 0.029 & 0.016  \\
& \textsc{BERTScore} & 0.055 & 0.018 & \textbf{0.026} & 0.017  \\
% &\textsc{Unsure}     & 29.85 & 29.71 & 34.53 & 25.30  \\
\bottomrule
\end{tabular}
}
\caption{\textbf{Correlation between Human Utility of Rationales and Task Performance/BERTScore:} Shown here are the correlation scores between task performance/BERTScore and Human Utility for T5-Large, T5-3B and davinci-instruct-beta(GPT-3). We use Theill's $U$ for Task Performance and Correlation Ration $\eta$ for BERTScore \cite{bert-score}.}
\label{tab:correlation}
\vspace{-0.1cm}
\end{table}

\begin{table*}[h!]
\vspace{-0.8cm}
\centering
\scalebox{0.69}{
\begin{tabular}{lll}
\toprule
    \textbf{Original Question, Gold Rationale and Label} & \textbf{Generalization Question and Label} & \textbf{Generalization Type}  \\\midrule
    \makecell[l]{
     \colorbox{lblue} {\textit{Q:}} Was Iggy Pop named after his father?\\
    \colorbox{lpink}{\textit{R:}} Iggy Pop's birth name was James Newell Osterberg Jr. \\ The father of Iggy Pop was James Newell Osterberg Sr.\\
    \colorbox{lyellow}{\textit{A:}} Yes
    } & \makecell[l]{
    \colorbox{lblue} { \textit{Q:}} Was Iggy Pop's name derived from his father?\\
    \colorbox{lyellow}{\textit{A:}} Yes}
    &  \makecell[c]{\colorbox{lpurple}{\textsc{Rephrase}}}\\ 
    \midrule
        \makecell[l]{
    \colorbox{lblue} { \textit{Q:}} Can the Moscow Kremlin fit inside Disney Land?\\
   \colorbox{lpink}{ \textit{R:}} The Moscow Kremlin is a fortified complex in the \\ 
    middle of Moscow Russia. The Kremlin takes up sixty \\
    eight acres. Disney Land is an amusement park in California. \\
    Disney Land occupies eighty five acres.\\
    \colorbox{lyellow}{\textit{A:}} Yes
    } & \makecell[l]{
    \colorbox{lblue} { \textit{Q:}} Is the Moscow Kremlin bigger than Disney Land?\\
   \colorbox{lyellow}{ \textit{A:}} No}
    &  \makecell[c]{\colorbox{lorange}{\textsc{Counterfactual}}}\\ 
    % \midrule
    %     \makecell[l]{
    % \textit{Q:} Does Julia Roberts lose the prolific acting contest in her family?\\
    % \textit{R}: As of May 2020, Julia Roberts has acted in 64 projects. \\
    % Julia Roberts has a brother in acting, Eric Roberts, \\
    % and a niece in acting, Emma Roberts. As of May 2020, \\
    % Eric Roberts has acted in 577 projects.\\
    % \textit{A:} Yes
    % } & \makecell[l]{
    % \textit{Q}: Does Julia Roberts have more acting projects than \\ 
    % her brother?\\
    % \textit{A}: No}
    % &  \makecell[c]{\colorbox{lorange}{Counterfactual}}\\ 
    % \midrule
    %     \makecell[l]{
    % \textit{Q:} Does Snoop Dogg advocate a straight edge lifestyle? \\
    % \textit{R}: A straight edge lifestyle requires abstaining from \\ 
    % the usage of recreational drugs or alcohol. Snoop Dogg is \\
    % famous for his chronic usage of marijuana.\\
    % \textit{A:} No
    % } & \makecell[l]{
    % \textit{Q}: Does Snoop Dogg advocate the use of recreational drugs \\
    % or alcohol?\\
    % \textit{A}: Yes}
    % &  \makecell[c]{\colorbox{lorange}{Counterfactual}}\\ 
    \midrule
        \makecell[l]{
   \colorbox{lblue} {\textit{Q:}} Can vitamin C rich fruits be bad for health?\\
    \colorbox{lpink}{\textit{R:} }Oranges are fruits that are rich in vitamin C. \\
    Oranges are very acidic fruits that can wear down tooth \\
    enamel. Too much Vitamin C can cause nausea and diarrhea.\\
    \colorbox{lyellow}{\textit{A:}} Yes
    } & \makecell[l]{
     \colorbox{lblue} {\textit{Q:}} Can oranges be bad for health?\\
    \colorbox{lyellow}{\textit{A:}} Yes}
    &  \makecell[c]{\colorbox{lgreen}{\textsc{Similar Reasoning}}}\\ 
    % \midrule
    %     \makecell[l]{
    % \textit{Q:} Is the Matrix a standalone movie?\\
    % \textit{R}: The Matrix ends in a cliffhanger. The story \\
    % is then resolved in two sequels, making a trilogy. \\
    % There are also supplemental works adding to the story, \\
    % such as a video game and the Animatrix.\\
    % \textit{A:} No
    % } & \makecell[l]{
    % \textit{Q}: Is the Matrix a trilogy?\\
    % \textit{A}: Yes}
    % &  \makecell[c]{\colorbox{lgreen}{Similar Reasoning}}\\ 
    % \midrule
    %     \makecell[l]{
    % \textit{Q:} Does water have viscosity?\\
    % \textit{R}: Viscosity is resistance of fluid to \\
    % deformation. Water is not resistant to deformation.\\
    % \textit{A:} No
    % } & \makecell[l]{
    % \textit{Q}: Is water resistant to deformation?\\
    % \textit{A}: No}
    % &  \makecell[c]{\colorbox{lgreen}{Similar Reasoning}}\\ 
    % \midrule
\bottomrule
\end{tabular}}
\caption{\textbf{Examples of generalization questions of each type from the StrategyQA Dataset}: We show the original question, rationale and label triplet, along with davinci-instruct-beta (GPT-3) generated generalization questions and gold label for the generated question.
% \xiang{color-code Q/A/R}
% \xiang{Do you need this in main text if we have Fig.~\ref{fig:generalization_pipeline}?}
}
\label{tab:generalization_examples}
\vspace{-0.4cm}
\end{table*}
% For each instance, we pick the majority vote \sahana{majority vote of what?} as the final label. 

% \begin{table}[h!]
% \scalebox{0.65}{
% \begin{tabular}{clcccc}
% \toprule
% & &\multicolumn{4}{c}{Correlation b/w metric and human utility} \\
% \cmidrule{3-6}
% Dataset & Type & All & GPT-3   & T5-3B   & T5-Large \\
% \midrule
% \multirow{2}{*}{StrategyQA} &\textsc{Task Accuracy} & 17.83 & 20.30 & 18.12 & 15.06  \\
% &\textsc{Similarity}     & 47.16 & 53.93 & 46.72 & 40.82  \\
% \midrule
% \multirow{2}{*}{OBQA} &\textsc{Task Accuracy}& 15.26 & 16.06 & 14.85 & 14.85  \\
% &\textsc{Similarity}     & 29.85 & 29.71 & 34.53 & 25.30  \\
% \bottomrule
% \end{tabular}
% }
% \caption{\textbf{Distribution of Human Utility of Rationales:} Shown here are the \%s of different types of rationales based on their utility, for different LMs and the two datasets.}
% \label{tab:correlations}
% \end{table}

Overall, while including annotations for all models combined, we observe that the correlation between task accuracy (whether a given instance was correctly predicted by the self-rationalizing model) and human utility of a rationale (useful, not useful and unsure) was close to none (Theill's $U = 0.0359$ and $U = 0.0221$ for StrategyQA and OBQA respectively). 
This indicates that while generating rationales might improve overall task performance, there is no guarantee that these rationales are useful for humans in solving the task correctly. 

In fact, if we look at the correlations for each LM separately, we observe Theill's $U$ for GPT-3, T5-3B and T5-Large were $0.111$ ($0.092$), $0.034$ ($0.029$) and $0.005$ ($0.016$) for StrategyQA (OBQA) respectively (Table \ref{tab:correlation}). 
This also demonstrates that even though T5-Large, which was fine-tuned on the entire training set had the highest task performance, it has the lowest correlation with human utility. 
% \brihi{Mention task performance of the 3 LMs early in the setup}.

We also compute the similarity between rationales and their corresponding gold rationale using BERTScore \cite{bert-score} for the test set, and compute their correlation with their human utility (Table \ref{tab:correlation}). 
For StrategyQA, the Correlation Ratio $\eta = 0.041$ for all three LMs combined, and $\eta=0.021, 0.017, 0.002$ for GPT-3, T5-3B and T5-Large respectively, whereas for OBQA $\eta = 0.055$ for all three LMs combined, and $\eta=0.018, 0.026, 0.017$ for GPT-3, T5-3B and T5-Large respectively.

\paragraph{What rationale properties are associated with human utility of rationales?}

% \xiang{It would be cool to list top-3/5 combination and their scores. maybe using acronyms on property names can save some space?}
We conduct a case-study for the StrategyQA dataset.
We list a set of desirable properties of that useful rationales should satisfy \cite{wiegreffe-etal-2021-measuring, wiegreffe-etal-2022-reframing, roscoe}.
These properties evaluate rationales along four axes - surface form qualities, support towards predicted labels, informativeness and style. 
Surface form qualities test whether a rationale is \textit{grammatical} and \textit{factually valid}. 
\textit{Association} with label and \textit{contrast} between different labels measure the extent to which rationales support the labels that were generated with them.
We also evaluate the informativeness of a rationale, which is determined by \textit{novel information} that the rationale provides over the question, along with asking whether it directly \textit{leaks the answer}.
Lastly, we also check whether the rationale contains \textit{irrelevant hallucinations} or relevant but \textit{redundant information}.
Descriptions and examples of these properties are shown in detail in Figure \ref{fig:properties}.

We use a Generalized Linear Mixed-Effects Model (GLMEM) (similar to \citet{lamm2020qed}) to estimate the importance of different properties and their interactions in predicting the human utility of rationales.
We observe that while in isolation or pairs, these properties are not sufficient indicators of human utility (\textsection \ref{sec:apx:prop_utility}), when all possible combinations of properties are considered, presence of all but coherence and association leads to a positive log odds for rationale utility: $0.139$. 
This implies that humans are generally robust to hallucinations that are irrelevant to the question.
Furthermore, association of the rationale with its predicted label is also not an important property for rationale utility, as the rationale may not be associated with the correct answer and therefore, mislead the human into making an incorrect choice.

% \section{Generalizing to New Questions with Rationales}
\section{Measuring Rationale Utility by Answering Generalization Questions}
\label{sec:generalization}

\begin{figure*}[h!]
\vspace{-0.8cm}
    \centering
    \includegraphics[width=\linewidth]{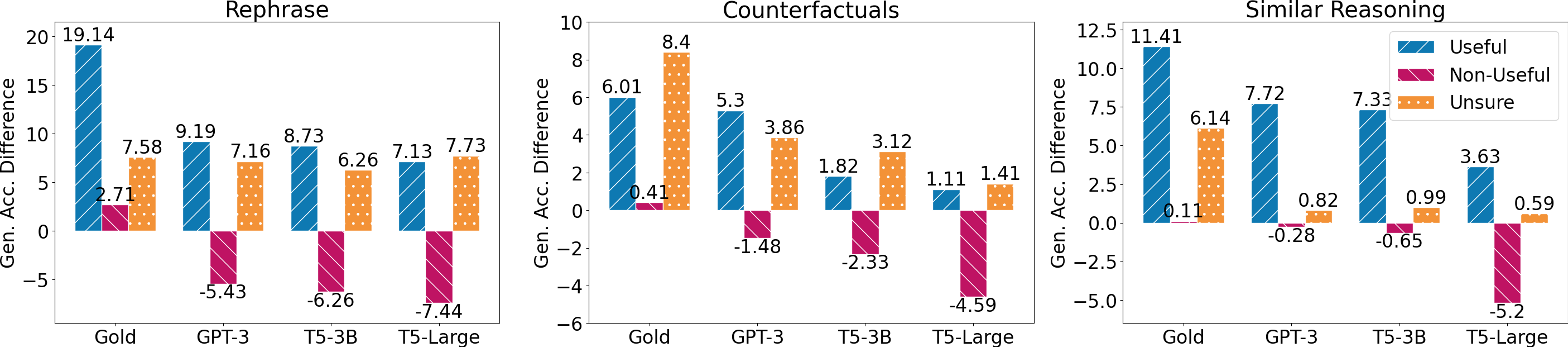}
    \caption{\textbf{Generalization Accuracy Difference for the StrategyQA Dataset:} In this Figure, we plot the \textit{difference} in accuracy of generalization questions, after and before a human annotator is shown the original question's rationale.}
    \label{fig:generalization_results}
    \vspace{-0.3cm}
\end{figure*}

As defined in \textsection \ref{sec:utility_setup}, human utility of rationales is determined by their ability to guide humans to correctly solve the task (instances).
We follow this up by investigating if humans can generalize to syntactic or semantic perturbations of the original question, while being shown rationales of the original question.
This will help us understand if human utility of rationales can also indicate whether rationales help with knowledge transfer for unseen instances.
For all our experiments, we use the StrategyQA Dataset.
% \brihi{Add intro paragraph}.
% \brihi{Need to find citations for human generalization to new instances}.
% \sahana{Can you add a line saying what you mean by `Useful' rationales? Do you mean the rationales which you found had human utility in Section 3?}
% \sahana{Do you mean that the new instances use the same rationale (i.e., the rationale that was found to have utility)? I think you should clarify why this section is relevant.}
\paragraph{Types of Generalization Questions.}

For our study, we consider three distinct types of generalization setups. 
Firstly, we evaluate the human $\mathcal{H}$'s ability to generalize to non-trivial \textbf{rephrases} of the original question.
% These instances are non-trivial rephrases of the original question. 
We avoid simple rephrases like changing a preposition, or removing an adverb so as to avoid near duplicates of the original question.
Next, we look at \textbf{counterfactual} questions. 
These questions follow the same reasoning steps as the original question, however, they flip the answer of the original question. 
Lastly, we test $\mathcal{H}$'s ability to understand questions that follow a \textbf{similar reasoning} process as the original question, but are not related to the original question.
These questions can entail entity swaps, or questions that use one of the reasoning steps to answer the original question.
% \brihi{Add generalization examples table}
Examples of each type of generalization question is shown in Table \ref{tab:generalization_examples}.

\paragraph{Generating Generalization Questions.} 

% \begin{figure*}[h!]
%     \centering
%     \includegraphics[width=\linewidth]{figures/generalization.png}
%     \caption{\textbf{Generating Generalization Questions:} A question, rationale, answer triplet is fed to GPT-3 using given templates. GPT-3 generates potential generalization question candidates from each template, which are then manually filtered and answered to create gold labels in-house.
%     \xiang{Probably need to compress for saving space. We can minimally just show the questions alone, and leave others to appendix.}
%     }
%     \label{fig:generalization_pipeline}
% \end{figure*}

For generating generalization questions as described above, we follow the Human and AI collaboration paradigm for dataset collection as introduced by \citet{wanli}.
% \brihi{Confirm number of demonstrations and types for each category from Ziyi.}
We first start by manually creating templates with instructions for each type of generalization question.
We then select six demonstrations for these templates.
The selected instructions and demonstrations are in Appendix (Table \ref{tab:generalization_demonstrations}).
These demonstrations are fixed for each type (however, may differ across the different types) and are selected from the training set.
% \brihi{Add figure ref}
For every test instance, we insert it at the end of the corresponding template, which is then used as a prompt for GPT-3 to generate questions.
To increase the number of good-quality generalization questions, we use GPT-3 to generate 5 generalization questions of each type for a given question, along with their answers.
We also vary the temperature ($0.7$) to control for diversity in generated questions.
% We also conducted a round of hyperparameter search of temperature which controls how much randomness is in the output on a small set of 50 instances. 
% We found that setting temperature = 0.7 performed the best and we used this setting for the whole generation.
The generated questions and their answers are then validated by a human study, to make sure that the final set of questions is of good quality (Details in \textsection \ref{sec:turk_gen_question}).

In the end, for each original question in the StrategyQA dataset, we obtain generalization questions of three different types, although the number of generalization questions per original question can vary. 
Overall, we collected $9659$, $1164$ and $2608$ generalization questions for the training, validation and test set, with $5.86$, $6.32$ and $5.70$ generalization questions per original question on average, respectively.
% We then manually filter the generated questions in-house and provide gold answers for them, to make sure that the final set of questions are of good quality.
% Finally, we have a set of $123$ rephrase, $102$ counterfactual and $171$ similar reasoning generalization questions that are of good quality from the test set.
% \brihi{Add details about \# of each type}
% The pipeline and examples of the templates are shown in Figure \ref{fig:generalization_pipeline}.

\begin{figure*}[h!]
\centering
\vspace{-1.1cm}
\hspace{-0.7cm}
    \includegraphics[width=0.88\linewidth]{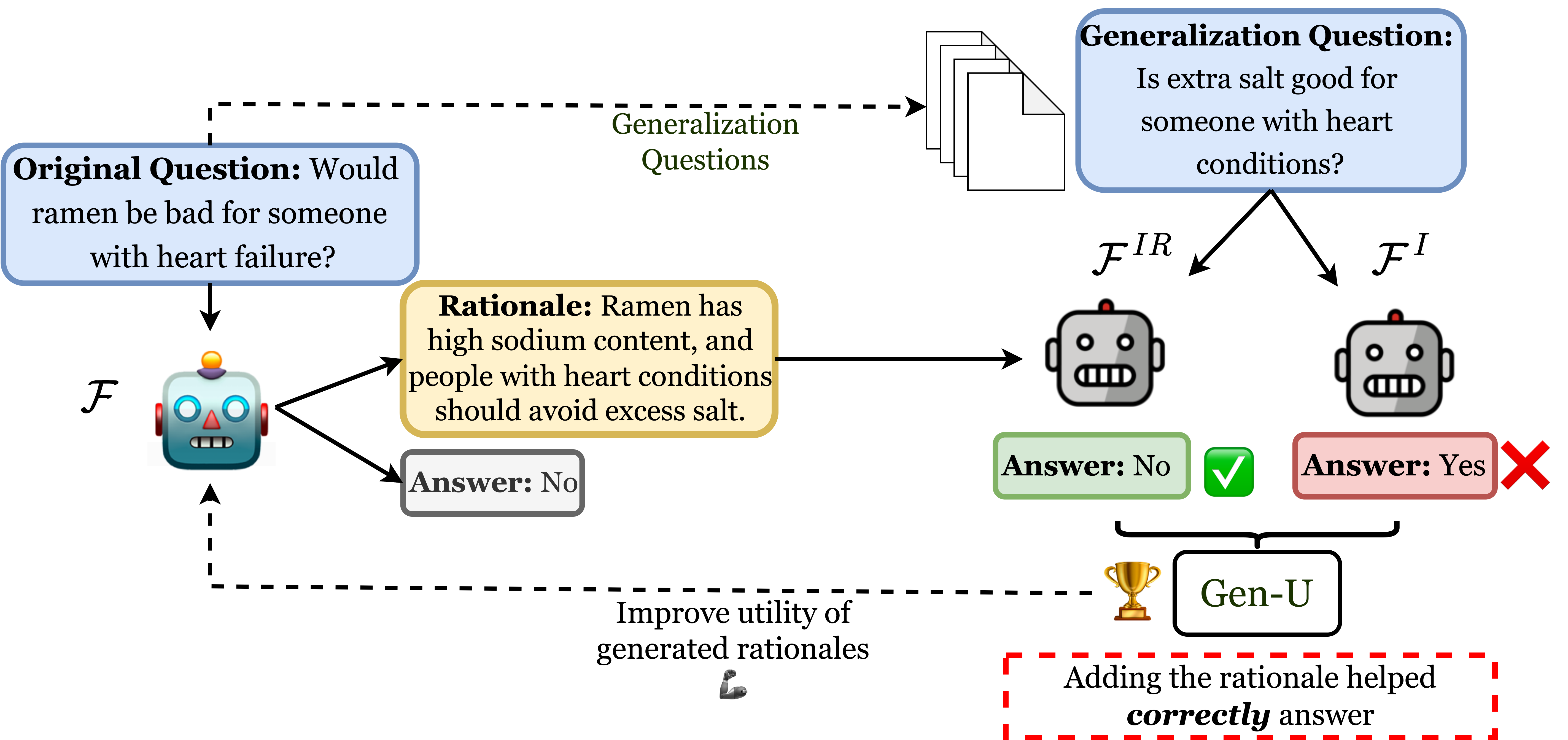}
    \caption{\textbf{Updating self-rationalising LMs with \reward:} Based on the generalization ability of two other LMs, we use \reward to update $\mathcal{F}$, so as to generate rationales with better utility.
    % \xiang{adjust figure to reduce vertical space.}
    }
    \label{fig:quark_figure}
    \vspace{-0.4cm}
\end{figure*}

\paragraph{Human generalization is a good indicator of human utility.} 

Similar to \textsection \ref{sec:utility_setup}, we first ask the annotators to answer a generalization question without the rationale.
We then show them the rationale of the original question, and ask them to answer the generalization question again, taking the rationale into account.
% Similar to \textsection \ref{sec:perf_util}, we first ask the annotators to answer a given question with and without the rationale.
% We then show them a generalization question, which they have to answer while referencing the rationale for the original question. 
We repeat the experiment above with rationales from the three LMs, along with gold rationales.
Each instance is annotated by five annotators.
Given that there are no corresponding rationales for the generalization questions, this annotation setup would measure the impact of rationales of the original question towards answering the generalization questions.
% \sahana{Ok, you have answered my previous note here. It might be worth adding a small note about this over there as well?}
% \paragraph{Results.}

% \brihi{Improve bar plot with larger text and number on them, and talk "numerically" in this text.}
In Figure \ref{fig:generalization_results}, we plot the difference between the generalization accuracies after and before being shown the rationale of the original question.
We observe that gold rationales form an upper bound for generalization difference, across all types of generalization questions and types of rationale utility.
Useful rationales are able to help humans generalize better to new instances, whereas non-useful rationales often \textit{mislead} humans to make incorrect choices, who might have correctly answered the question before, which is indicated by the \textit{negative} plot bars in the Figure.
Rationales about which we are unsure are better or close to useful rationales for rephrase and counterfactuals, as these generalization questions are relatively simpler.

However, for similar reasoning questions, they underperform useful rationales.
This indicates that for rationales that are unsure, either the human was already aware of the answer or the questions are easier to answer as humans are able to answer rephrases and counterfactuals correctly, but fail in generalizing to questions that follow a similar reasoning process.
We can also note that GPT-3 generated rationales help generalize better to more difficult settings like counterfactuals or similar reasoning questions.
% Figure \ref{fig:generalization_results} shows generalization accuracy of rationales with different human utility, split across different LMs generating them as well as gold rationales.
% We observe that gold rationales form an upper bound for generalization accuracy, across all types of generalization questions and types of rationale utility.
% Useful rationales are always significantly better than non-useful rationales while answering generalization questions, which indicates that rationale utility is a strong signal for generalizing to newer instances.
% Paraphrases of original questions are relatively easier to answer, when compared to counterfactuals and similar reasoning, as it can be seen by both Gold rationale, as well as accuracy of all models combined.
% We can also note that GPT-3 generated rationales help generalize better to more difficult scenarios like counterfactuals or similar reasoning questions.
% Rationales that we are unsure of in terms of utility are helpful in generalizing to paraphrases or similar reasoning questions.
% This indicates the presence of easy-to-answer questions in that bucket of instances, which also hints towards the fact that their paraphrases or similar reasoning counterparts would also be easier to answer. 
% \sahana{Can you rephrase this sentence? I don't understand it.}
% Counterfactuals however show a different trend, where unsure rationales have a lower generalization accuracy when compared to useful rationales. 
Examples of generalization questions that were answered correctly/incorrectly for rationales that have high or low human utility is shown in the Appendix (Table \ref{tab:generalization_utility_examples}).
% \paragraph{}

\section{Improving Human Utility of Self-Rationalising LMs}
\label{sec:genlm}

Smaller LMs like T5-large have better task accuracy, but lack in generating more useful rationales.
It can be observed (\textsection \ref{sec:utility_setup}) that the task performance of a self-rationalizing LM and the human utility of its corresponding generated rationales are not correlated.
Based on our insights about how useful rationales can help humans generalization to unseen questions, we propose \reward, which simulates a human through an LM: we define and use \reward to improve human utility of smaller LMs like T5-large, while aiming to maintain their task accuracy (Figure \ref{fig:quark_figure}).
For all our experiments, we use the StrategyQA Dataset.
% Intro line - in section 2 we discussed what human utility is, and explored various rationale/LM properties and their correlation to tis human utilty. Then in section 3, we introduced and described a method to \textit{quantitatively measure} the human utility of a rationale; namely, by measuring human generalization of the rationale. In this section we work on \textit{improving} the human utility of the rationales generated by a language model, by first creating an automatic metric that will compute the degree of generalization and then using this metric to iteratively train the model to generate rationales of higher utility using Quark \cite{lu2022quark}.

% Things to write - (1) metric definition (2) experimental details (hparams go in appendix??) (3) the improvement we got in both automatic metric and subsequent human analysis (4) talk a bit about how the utility in terms of gen-U was low at the beginning itself and how it's better now, but the low number is again an indicator that LMs don't give really useful rationales (5) possible future extensions of this section (maybe get even more gen qns etc) to make this improvement stronger

\paragraph{LM generalization is a better indicator of rationale's human utility. }

\textsection \ref{sec:generalization} indicated that generalization to unseen but similar questions via rationales of the original question is a reasonable proxy for human utility of rationales.
Based on this insight, we propose \reward, which estimates the generalization performance of an LM variant, after and before being shown a rationale generated by a self-rationalizing model.

For a given input-output pair $x,y$, there exist a set of $n$ generalization questions $X_{g}, Y_{g} = \{(x_{g1}, y_{g1}), (x_{g2}, y_{g2}), \dots, (x_{gn}, y_{gn})\}$ that is created as per \textsection \ref{sec:generalization}.
Let $\mathcal{F}$ be a self-rationalising LM as defined in \textsection \ref{sec:utility_setup}, for which we want to estimate the score. 
Let $\mathcal{F^{I}}$ be an LM that takes in $X_{g}$ as its input and predicts a set of labels $Y^{I}_{g}$.
Similarly, $\mathcal{F^{IR}}$ be an LM that takes in $X_{g}$ and the rationale $r_{p}$ generated for $x$ by $\mathcal{F}$, and predicts a set of labels $Y^{IR}_{g}$.

\reward for $x$ is defined as:
% $$
% \resizebox{.9\hsize}{!}{
%  \textsc{mode}_{i=1:n}\bigg(
% \begin{cases} 
% \left(1-\mathds{1}(y^{I}_{gi} = y_{gi})\right) & y^{IR}_{gi} = y_{gi} \\
% -1 & y^{IR}_{gi} \neq y_{gi}
% \end{cases}
% \bigg) }
% $$

$$
\textsc{mode}_{i=1:n}\bigg(
\begin{cases} 
\left(1-\mathds{1}(y^{I}_{gi} = y_{gi})\right) & y^{IR}_{gi} = y_{gi} \\
-1 & y^{IR}_{gi} \neq y_{gi}
\end{cases}
\bigg)
$$

Here, $\textsc{mode}$ returns the most frequently occurring value from the set (similar to majority voting in a set).
In other words, if a generalization question is answered incorrectly after being shown the rationale, \reward is $-1$, otherwise, \reward calibrates itself w.r.t the answer before being shown the rationale, to accommodate for cases where the question is easy-to-answer or the LM already contains relevant background knowledge.
Then, we pick the majority vote of the scores (depicted by the mode) for all the generalization questions for a given original question as its score.

To validate if \reward is indeed usable, we calculate correlations between \reward and human utility of the corresponding rationales.
We find that Theill's $U=0.22$, which is indicates that \reward is a better estimate that $\mathcal{F}$'s task accuracy or BERTScore similarity between generated and gold rationales (refer Table \ref{tab:genu_correlation} for correlation scores).
% \brihi{Add correlation scores}

\begin{table}[h]
    \centering
    \scalebox{0.79}{
    \begin{tabular}{c|ccc}
    \hline
       Metric  & \reward & \textsc{Task Accuracy} & \textsc{BERTScore} \\ \hline
       Correlation  & 0.227 & 0.035 & 0.041 \\ \hline
    \end{tabular}}
    \caption{\textbf{Improvement in Correlation Scores for the StrategyQA Dataset:} We observe that \reward leads to a better correlation with human utility than Task Accuracy or BERTScore.}
    \label{tab:genu_correlation}
    \vspace{-0.3cm}
\end{table}

\paragraph{\reward as a reward for updating LM.}
We use the Quark \cite{lu2022quark} algorithm with \reward to improve the human utility of rationales generated by $\mathcal{F}$. 
Quark is an RL-inspired training algorithm that uses reward signals as control tokens on the encoder (or decoder) side, to condition the generation of text. 
% 6 \cite{meister-entropyreg} .

For $\mathcal{F}$, we use the same T5-large setup used in \textsection \ref{sec:utility_setup}. 
For implementing \reward, we use T5-base LMs for $\mathcal{F^{I}}$ and $\mathcal{F^{IR}}$, which are both finetuned on the StrategyQA dataset.
We begin by first fine-tuning $\mathcal{F}$ for $25$ epochs with supervised learning on the StrategyQA data, after which we continue training with Quark.
The final $\mathcal{F'}$ is obtained after finding the best hyperparameter choices based on \reward scores for the validation set. %\brihi{Sahana: Add these to the appendix.}
% We run experiments on the StrategyQA dataset with the t5-large I-OR infilling model (change name to what has been used before). 
% We first train the model with xx epochs of supervised training, after which we continue training with Quark. 
% We tried various hyperparameter settings for both Quark and \reward and chose the best performing model checkpoint based on the validation \reward scores. 

\begin{table}[t!]
\centering
\scalebox{0.8}{
\begin{tabular}{lcc|c}
\toprule
                       & $\mathcal{F}$ & $\mathcal{F'}$ (w/ Quark) & GPT-3-175B\\
\midrule
\reward                &   -0.315     &   -0.26 $\uparrow$  &   -        \\
Task Accuracy          &   67.03     &   65.06 $\downarrow$  &   60.04    \\
\% \textsc{Useful}     &   15.06     &   17.01 $\uparrow$  &   20.30    \\
\% \textsc{Not Useful} &   44.10     &   40.20 $\uparrow$  &   25.76    \\
\% \textsc{Unsure}     &   40.82     &   42.79 $\downarrow$  &   53.93    \\
\midrule
\# of Params           &   770M     &    770M    &   175B \\
\bottomrule
\end{tabular}
}
\caption{\textbf{Impact of \reward as a reward to update LM using Quark \cite{lu2022quark} algorithm:} On the StrategyQA Dataset, we show the \% of different types of rationales for the LM before ($\mathcal{F}$) and after ($\mathcal{F'}$) being updated with generation feedback through the Quark algorithm, using \reward as the reward. We also note the \% of rationales for 
davinci-instruct-beta (GPT-3), which is the best performing variant in terms of human utility. Here, $\uparrow$ implies improvement seen in $\mathcal{F'}$, and vice versa.}
\label{tab:quark_results}
\vspace{-0.3cm}
\end{table}
% For sahana's reference: The results filled in this table is model 40, checkpoint 5500

Table \ref{tab:quark_results} demonstrates the \reward scores before and after using Quark to update $\mathcal{F}$.
On the updated LM $\mathcal{F'}$, we conduct the same human utility evaluations as done in \textsection \ref{sec:utility_setup} to evaluate the improvement observed by lay humans.
We note that the updated LM is able to retain most of the task performance, while improving the \% of \textsc{Useful} rationales by $2\%$.
\reward also helps in getting rid of $4\%$ of mislead (\textsc{Not Useful}) rationales.
We also compare the updated LM with GPT-3, which yielded the best human utility of rationales.
\reward is able to make the updated LM closer to the human utility of GPT-3, while ensuring the task performance for the updated LM remains better than GPT-3.
This indicates that while incorporating human utility while generating rationales is a difficult problem and there is room for improvement, smaller LMs like T5-large are capable of improving, without compromising on the task accuracy that is obtained via fine-tuning.
% We run MTurk eval on the test set rationales generated by this best checkpoint, and we observed that the amount of useful rationales increased from $15.06\%$ to $17.01\%$.
% \xiang{Make a small table to show, for before vs. after Quark updates: task perf, \% useful rationales, \% non-useful, \% unsure.}
% \paragraph{Noting down hparams which should go into appendix}
% Base model: I-OR t5-large, with infilling template
% \reward I-O and IR-O models: t5-base trained with infilling template
% Quark learning rate: $1e-5$, gradient clipped at $1.0$, KL-divergence and entropy coefficients $0.05$
% Sampling: Two samples generated per training instance, every 500 steps, with top-p ($0.7$) sampling.

% \input{sections/2_background}
% \input{sections/3_rq1}
% \input{sections/4_rq2}
% \input{sections/5_rq_gen}
\section{Related Work}
\label{sec:related_work}

% \brihi{Check for missing references, headings seem repetitive so need to find a way to remove "FTR"}

\paragraph{Evaluating free-text rationales}

Extractive explanations have been used to improve human's understanding of the model \cite{wang2021areexplanations, feng2018whatcanai, carton2020feature, chen2022usecase, idahl-etal-2021-towards, chu2020arevisual} or detecting errors in model predictions \cite{gonzalez-etal-2021-explanations}.
Although prior motivation of generating rationales has been primarily to improve task model performance \cite{rajani-etal-2019-explain,star,cot,icl-lampinen}, recent works have evaluated rationales in various ways. 
\citet{wiegreffe-etal-2022-reframing} use human acceptability judgements on over-generated rationales by GPT-3 \cite{gpt3} where \citet{suninvestigating} measure benefits of rationales to LMs.

\paragraph{Updating LMs with Generation Feedback}
% \brihi{Quark or everything that is like quark will come here}
There are several ways to update language models with rewards to correct misaligned behaviour that models learn \cite{chen2021decision, janner2021offline}. 
\citet{lu2022quark} unlearn these misalignments by fine-tuning the language model on signals of what not to do. 
Similarly, \citet{star} iteratively leverage a small number of rationale examples to training and only keep good examples. 
% It is also popular to view the RL as a sequence modeling problem and output the optimal\cite{chen2021decision} or generate a sequence of actions that leads to a sequence of high rewards.\cite{janner2021offline}. 
Our method is inspired by several evaluation methods \cite{chen2022rev,chan2022frame,wiegreffe2020measuring,hase2020leakage} which discussed how to better evaluate the quality of free-text rationales with regard to labels and contexts.
\section{Conclusion and Future Work}
In this work, we study human utility of free-text rationales, by measuring how well lay humans are able to solve tasks with their help.
Through extensive human evaluations, we show that human utility of rationales generated by current LMs is rather unsatisfactory, and existing available measures do not correlate well with it.
We find that generalization ability with rationales as context is a good proxy for human utility, and use it as a reward to improve human utility of LMs.

There are a lot of scopes to improve human utility of self-rationalising LMs, where granular-level properties of rationales can be leveraged directly.
Furthermore, evaluation of human utility on other tasks (like closed-book QA) is something that is also worth looking at, given that human annotators cannot `guess' answers for these tasks, making it harder for LMs and humans alike.
\section{Acknowledgments}

This research is supported in part by the Office of the Director of National Intelligence (ODNI), Intelligence Advanced Research Projects Activity (IARPA), via Contract Nos. 2019-19051600007, 2022-22072200006, NSF IIS 2048211, and gift awards from Google, Amazon, JP Morgan, and Sony.
We would like to thank all of our collaborators at USC NLP Group, USC INK Research Lab, Meta AI and AI2, specially Swabha Swayamdipta and Ameya Godbole for their constructive feedback on this work.
\section*{Limitations}

\paragraph{Estimating human utility is expensive.}
The core of our work is built on conducting extensive human evaluations, to understand how well lay humans can solve tasks with rationales.
In order to replicate these findings to other tasks, one would require the same scale of human evaluations, which are expensive and tedious.
These tasks are also difficult to explain to lay crowdworkers, because of which several rounds of turking are required to reach good annotator agreements.
% It is also difficult to arrive at an acceptable annotator agreement, which we could obtain after several rounds of trials.
Given these shortcomings of human evaluation, a reliable metric that estimates human utility is necessary.
% \xiang{Add a sentence to pitch having reliable metric is important.}

\paragraph{Generating generalization questions is not completely automated.}
Even though we prompt GPT-3 with varied demonstrations to generate generalization questions of each type, we still have to manually filter them (via crowdsourcing) to obtain a cleaner set of questions.
Furthermore, in order to obtain gold answers of these questions, we generate answers by prompting GPT-3 again, which also requires further validation.
A completely automated method of generating these questions would lead LM updates to be independent of human involvement.

\paragraph{Even though \reward has a better correlation with human utility, the correlation is still low.}
To train models to produce free-text rationales with more human-utility through Quark \cite{lu2022quark}, it is first necessary to have an accurate metric that can serve as a reward function/scoring metric for human utility. 
In this work, we found that human generalization is good indicator of human-utility. 
However, given that Quark requires frequent reward scoring, it is infeasible to use human annotations for the same. 
Our proposed automatic metric \reward that simulates human generalization has a good correlation with human utility (better than task accuracy, or BERTScore), but overall, it still has a low correlation with human utility of rationales.
Developing a score with better correlation with human utility (perhaps even a stronger version of \reward) will decrease the effect of this limitation and lead to training that further increases human utility of generated rationales.

\section*{Ethics Statement}
\paragraph{Data. }All the datasets that we use in our work are released publicly for usage and have been duly attributed to their original authors.
Data for all human studies that we conduct will be publicly released with this work, with appropriate annotator anonymisations.
% We do not collect any datasets as a part of our work.

\paragraph{Crowdsourcing. } 
All our crowdworkers are from countries where English is the primary language.
For all our human studies, the task is setup in a manner that ensure that the annotators receive compensation that is above minimum wage (\$15/hour).
Since we conduct extensive qualification tasks before annotations, crowdworkers that participate in the qualification are compensated more than the task, given the time taken to read and understand task instructions and examples.
Furthermore, we ensure that we correspond with crowdworkers over email to address their queries.
Crowdworkers have also been given bonuses for flagging errors in the task, or consistently providing good-quality annotations.
% As part of our user study conducted in Section \ref{sec:exp:rq4}, we collected information about the time taken to annotate rationales and labels for instances. We provide the instructions given to MTurkers in Appendix \ref{sec:app:time_budget}, along with screenshots of the UI displayed to them. Further details about the task setup and results are provided in Section \ref{sec:exp:rq4}. 
% Each task is setup in a manner that ensure that the annotators receive compensation that is above minimum wage. Furthermore, we ensure that we correspond with MTurkers over email to address their queries regarding rejection of HITs, many of which have been overturned after thorough discussions. 

\bibliography{custom}
\bibliographystyle{acl_natbib}

\appendix

\newpage
\section{Appendix}
\label{sec:appendix}

\subsection{Extended Related Work}
Here, we discuss more about different techniques to generate free-text rationales from LMs, as well as touch upon some Psychology and Cognitive Science literature that insipired our study.

\paragraph{Rationale Generation}

There are two distinct methods of generating free-text rationales. 
The first way is to fine-tune an encoder-decoder like model, for example, T5 or it's variations like UnifiedQA \cite{raffel-T5, khashabi2022unifiedqa, 2020unifiedqa}. 
Finetuning T5 to generate rationales \cite{narang2020wt5, paranjape-etal-2021-prompting} entails appending a tag like \texttt{explain:} in the input text, to nudge the LM to generate rationales during prediction. 
The generated text can either be separated by structured tags like \texttt{answer:, explanation:}, or it can be unstructured, with the answer followed by a \texttt{because} keyword, followed by the rationale. 
Recent methods have also analysed few-shot prompting of T5 with different input-output templates \cite{marasovic-etal-2022-shot}. 
Another recent approach of generating free-text rationales is via in-context learning \cite{cot, kojima2022large, marasovic-etal-2022-shot, wiegreffe-etal-2022-reframing}. 
A decoder-only model like GPT-3 or its variants \cite{gpt3, gpt-j} that are pretrained on a larger corpora of world-knowledge are prompted with demonstrations \cite{cot}, wherein each example contains its corresponding explanation. 

\paragraph{Human Utility of Human Rationales}

Several works in Psychology and Cognitive Science detail the role that human rationales play for human understanding. 
These studies have shown that human rationales are inherently incomplete and do not capture the complete deductive reasoning process. \cite{chenhaolimits}. 
These rationales are used to either provide \textit{evidence} or \textit{procedure} behind obtaining a given conclusion for a situation \cite{Lombrozo2006-zi}. 
Furthermore, some works have also detailed the utility human rationales have for human understanding. 
Human rationales have shown to help better generalise to unknown circumstances \cite{explandinference}, justify decision-making \cite{motivatedexpl}, understand relationships between different world entities \cite{analogy_expl_proof}, diagnose when something went or might go wrong, as well as explain one off events that are bizarre \cite{Keil2006-yt}.

\subsection{Task and Dataset Selection}
\label{sec:apx:task_dataset_selection}

% \xiang{Want to be sure that we speak out about this caveat in main text and leave a pointer to here.}
We refrain from tasks used in existing free-text rationale works \cite{dataset_review} like NLI \cite{esnli} and Commonsense QA \cite{aggarwal-etal-2021-explanations}.
A primary reason for this is that humans are already able to reason better than models for NLI and Commonsense QA \cite{nangia2019muppet, talmor2021commonsenseqa}.
Therefore, the objective of machine rationales in this case is just to establish trust or generate faithful rationales. 
We aim to study rationale utility specifically in cases where the rationales can help with knowledge transfer that helps humans to correctly solve a task.
We thus impose the following constraints in our task and dataset selection:
\begin{itemize}
\setlength\itemsep{0em}
\item \textbf{Added advantage:} Tasks where machines can provide added advantage and that are not trivial or obvious for humans to solve.
\item \textbf{Objectivity:} Tasks where the reasoning has a limited scope of subjectivity.
% Tasks which are objective in nature, and where the reasoning has limited scope of subjectivity \sahana{What does `subjective' mean?}.
\item \textbf{Dataset size (of rationale annotations}): Size of gold rationales is considerably larger in the dataset, so as to provide room for training LMs with those rationales.
% \sahana{Do you mean to say dataset should have gold rationales?}.
\end{itemize}
In this work, we choose the StrategyQA dataset \cite{geva-etal-2021-aristotle}, which is an open-domain binary QA benchmark, where questions require implicit reasoning steps to be answered.
The StrategyQA dataset consists of an input question, the answer, along with intermediate implicit reasoning steps that are used to answer the questions. 
The implicit reasoning steps were generated by decomposing the original question into multiple questions. 
For our project, we combine these implicit reasoning steps and use them as rationales for a given instance. 
We also use the OpenBookQA Dataset \cite{Mihaylov2018CanAS} for validating human utility of rationales for existing LMs.
Both of these datasets are available publicly for use, and have been checked manually by authors for toxic/offensive content.

\begin{table}[h!]
\centering
\scalebox{0.85}{
\begin{tabular} {c|ccc}
\toprule
Split & Train & Dev & Test \\
\midrule
Number & 1648 & 184 & 458\\
\bottomrule
\end{tabular}
}
\caption{\textbf{Dataset details}: Since the original test set of StrategyQA does not have gold labels, we used only the original train set and validation set in our experiments. Our test set is the original validation set, and our train and validation sets are splits ($90/10\%$) from the original train set. }
\label{tab:dataset_details}
\end{table}

\begin{figure*}[h!]
    \centering
    \includegraphics[width=\linewidth]{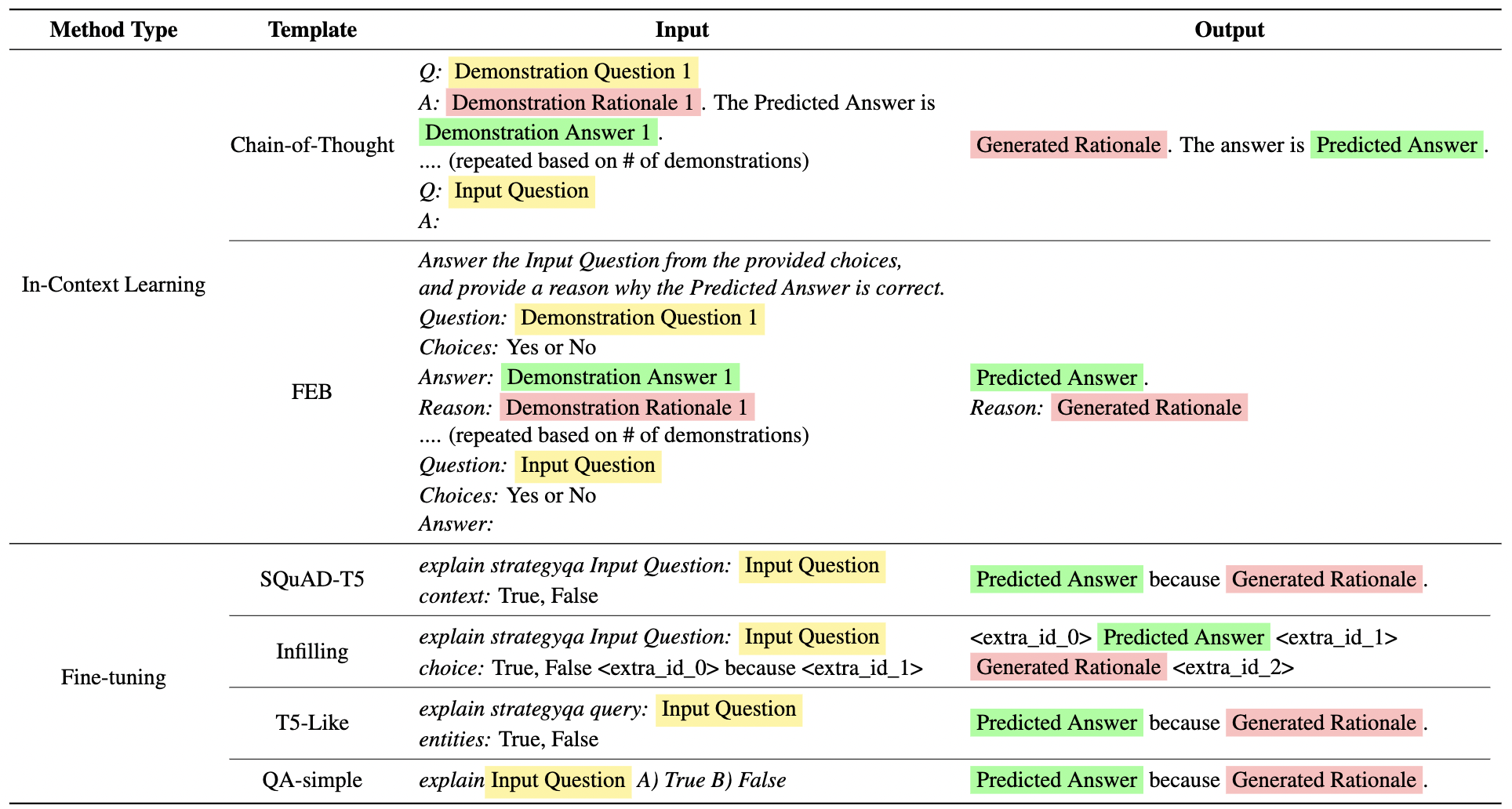}
    \caption{\textbf{Prompt templates for generating rationales:} Shown here are inputs and outputs of different template variations. Chain-of-Thought templates are taken from publicly released versions by \citet{cot}, whereas FEB and Fine-tuning templates are taken from \citet{marasovic-etal-2022-shot}.}
    \label{fig:prompt_template}
\end{figure*}

\subsection{Self-Rationalising Models}
\label{sec:apx:self_rationalising_models}

\begin{table*}[h!]
\centering
\scalebox{0.80}{
\begin{tabular}{cccccccc}
 \toprule
\textbf{} & \textbf{} & \textbf{}&\textbf{} & \multicolumn{4}{c}{\textbf{Accuracy}} \\
\cmidrule{5-8}
\textbf{$\mathcal{F}$}&\textbf{Model} & \textbf{Size} & \textbf{Finetuning setting} & \textbf{SQuAD-T5} & \textbf{Infilling} & \textbf{QA-simple} & \textbf{T5-like} \\
 \midrule
\multirow{6}{*}{Without Rationale}&\multirow{3}{*}{StrategyQA} & large & full & \cellcolor{lblue} 64.41 & 62.45 & 61.35 & 62.45 \\
 & & 3B & 48-shot & \cellcolor{lblue} 55.46 $\pm$ 3.47 & 53.35 $\pm$ 2.95 & 50.95 $\pm$ 3.85 & 52.84 $\pm$ 4.51 \\
 & & 3B & 128-shot & 60.48 $\pm$ 0.87 & 60.11 $\pm$ 2.21 & 52.47 $\pm$ 2.21 & \cellcolor{lblue} 61.50 $\pm$ 2.55 \\
%  \cmidrule{2-8}
% &
% \multirow{3}{*}{UnifiedQA} & large & full & 63.76 & 61.57 & 67.90 & \cellcolor{lblue} 68.34 \\
%  & & 3B & 48-shot & 54.80 $\pm$ 3.64 & 55.46 $\pm$ 4.36 & \cellcolor{lblue} 55.97 $\pm$ 3.01 & 55.24 $\pm$ 4.22 \\
%  & & 3B & 128-shot & 60.05 $\pm$ 3.08 & 58.22 $\pm$ 0.55 & \cellcolor{lblue} 61.50 $\pm$ 0.55 & 59.24 $\pm$ 5.18 \\
 \cmidrule{2-8}
&
\multirow{3}{*}{OBQA} & large & full & 71 & 65.8 & 69 & 70 \\
 & & 3B & 48-shot & 64.33 $\pm$ 2.30 & 61.87 $\pm$ 3.01 & 68.40 $\pm$ 0.69 & 63.93 $\pm$ 3.63 \\
 & & 3B & 128-shot & 68.27 $\pm$ 4.12 & 67.27 $\pm$ 1.53 & 71.20 $\pm$ 2.11 & 67.13 $\pm$ 0.42 \\
\midrule

 \multirow{6}{*}{With Rationale}&\multirow{3}{*}{StrategyQA} & large & full & 61.14 &\cellcolor{lred} 67.03 & 62.45 & 60.26 \\
& & 3B & 48-shot & 51.97 $\pm$ 1.00 & \cellcolor{lblue} 53.35 $\pm$ 1.33 & 50.94 $\pm$ 2.62 & 50.87 $\pm$ 3.28 \\
 & & 3B & 128-shot & 52.40 $\pm$ 2.19 & \cellcolor{lred} 56.70 $\pm$ 1.85 & 53.93 $\pm$ 3.61 & 53.35 $\pm$ 1.40 \\
%  \cmidrule{2-8}
%  & \multirow{3}{*}{UnifiedQA} & large & full & 64.85 & \cellcolor{lblue}65.72 & 62.45 & 62.45 \\
% & & 3B & 48-shot & 53.49 $\pm$ 4.36 & \cellcolor{lblue}60.99 $\pm$ 2.56 & 55.38 $\pm$ 5.70 & 55.09 $\pm$ 4.63 \\
%  & & 3B & 128-shot & 58.23 $\pm$ 3.07 & \cellcolor{lblue}62.08 $\pm$ 0.77 & 59.97 $\pm$ 4.94 & 57.50 $\pm$ 1.28 \\
 \cmidrule{2-8}
 & \multirow{3}{*}{OBQA} & large & full & 70.20 & 70.20 & 67.20 & 70.40 \\
& & 3B & 48-shot & 62.67 $\pm$ 2.34 & 63.07 $\pm$ 2.72 & 67.93 $\pm$ 4.84 & 66.60 $\pm$ 1.64 \\
 & & 3B & 128-shot & 67.47 $\pm$ 3.16 & 66.07 $\pm$ 2.66 & 70.40 $\pm$ 2.31 & 69.00 $\pm$ 0.53 \\

 \bottomrule
\end{tabular}
}
\caption{\textbf{Self-Rationalising Model Results (Fine-tuning)}: Shown here are test set accuracies of LMs (T5) of different sizes (large and 3B), and fine-tuned with different number of training examples, for four different templates. Cells highlighted in  \colorbox{lblue}{blue} are highest performing templates for each model configuration and \colorbox{lred}{red} denotes a configuration selected for the rest of our work.}
\label{tab:finetune_res}
\end{table*}

We try variations of in-context learning based approaches \cite{cot}, as well as few-shot and full finetuning approaches \cite{marasovic-etal-2022-shot} to generate rationales. 
For in-context learning based approaches, we vary the demonstrations based on the number of demonstrations desired, and the selection strategy for these demonstrations.
These demonstrations can either be fixed across all instances vs. randomly picked for each instance, from the training set.
Demonstrations that are picked randomly can either be six in number (to match a fixed number of demonstrations as per \citet{cot}), or determined by a maximum token length that is specific beforehand (for our experiments, we use $2048$ as the maximum token length of an input).
For these settings, we implement two input-output templates -- where rationales $r_p$ come after (FEB) \cite{marasovic-etal-2022-shot} or before the prediction $y_{hr}$ respectively (Chain-of-Thought or CoT) \cite{cot}. 
The LM used for all in-context learning experiments is GPT-3 \cite{gpt3}.
For fine-tuning approaches, we fine-tune two LMs - T5 \cite{t5} and UnifiedQA \cite{khashabi-etal-2020-unifiedqa}, with varying sizes - large and 3B. 
For each of these two LMs, we use four variations of input-output templates (SQuAD-T5, Infilling, T5-Like and QA-simple), as defined by \citet{marasovic-etal-2022-shot}. 
Examples of each of these templates are provided in Figure \ref{fig:prompt_template}. 

\begin{table}[h!]
\centering
\scalebox{0.73}{
\begin{tabular}{ccccc}
\toprule
\textbf{$\mathcal{F}$} & \textbf{Template} & \textbf{\# of demo} & \textbf{Demo Picked} & \textbf{Accuracy} \\
\midrule
\multirow{7}{*}{\makecell[c]{Without\\ Rationale}} & \multirow{3}{*}{CoT} & 6 & Randomly & \cellcolor{lblue} 57.11 \\
 & & max len & Randomly & 53.98 \\
 & & 6 & Fixed & 56.23 \\
 \cmidrule{2-5}
& \multirow{3}{*}{FEB} & 6 & Randomly & 52.84 \\
 & & max len & Randomly & 56.33 \\
 & & 6 & Fixed & 54.80 \\
\midrule
\midrule
\multirow{7}{*}{\makecell[c]{With\\ Rationale}} & \multirow{3}{*}{CoT} & 6 & Randomly & 58.51 \\
 & & max len & Randomly & 55.24 \\
 & & 6 & Fixed & 58.90 \\
 \cmidrule{2-5}
& \multirow{3}{*}{FEB} & 6 & Randomly & \cellcolor{lred} 60.04 \\
 & & max len & Randomly & \cellcolor{lblue} 60.04 \\
 & & 6 & Fixed & 57.42 \\
\bottomrule
\end{tabular}}
\caption{\textbf{Self-Rationalising Model Results (In-Context Learning) for StrategyQA Dataset}: Shown here are test set accuracies of davinci-instruct-beta (GPT-3), when it is prompted to predict with/without generating rationales. Cells highlighted in \colorbox{lblue}{blue} are highest performing variations, and \colorbox{lred}{red} denotes a configuration selected for the rest of our work.}
% \vspace{-1cm}
\label{tab:icl_res}
\end{table}

% $\spadesuit$
\begin{table}[h!]
\centering
\scalebox{0.73}{
\begin{tabular}{ccccc}
\toprule
\textbf{$\mathcal{F}$} & \textbf{Template} & \textbf{\# of demo} & \textbf{Demo Picked} & \textbf{Accuracy} \\
\midrule
\multirow{7}{*}{\makecell[c]{Without\\ Rationale}} & \multirow{3}{*}{CoT} & 6 & Randomly & \cellcolor{lblue} 57.11 \\
 & & max len & Randomly & 53.98 \\
 & & 6 & Fixed & 56.23 \\
 \cmidrule{2-5}
& \multirow{3}{*}{FEB} & 6 & Randomly & 52.84 \\
 & & max len & Randomly & 56.33 \\
 & & 6 & Fixed & 54.80 \\
\midrule
\midrule
\multirow{7}{*}{\makecell[c]{With\\ Rationale}} & \multirow{3}{*}{CoT} & 6 & Randomly & 53.60 \\
 & & max len & Randomly &  \cellcolor{lred} 55.60 \\
 \cmidrule{2-5}
& \multirow{3}{*}{FEB} & 6 & Randomly & 40.40 \\
 & & max len & Randomly & 41.20 \\
\bottomrule
\end{tabular}}
\caption{\textbf{Self-Rationalising Model Results (In-Context Learning) for OBQA Dataset}: Shown here are test set accuracies of davinci-instruct-beta (GPT-3), when it is prompted to predict with/without generating rationales. Cells highlighted in \colorbox{lblue}{blue} are highest performing variations, and \colorbox{lred}{red} denotes a configuration selected for the rest of our work.}
% \vspace{-1cm}
\label{tab:icl_res_obqa}
\end{table}

% $\spadesuit$

As seen in Tables \ref{tab:finetune_res}, \ref{tab:icl_res} and \ref{tab:icl_res_obqa}, for the StrategyQA and OBQA datasets, FEB templates with randomly selected demonstrations provides the highest accuracy for in-context learning approaches, whereas the infilling template consistently outperforms other input-output templates for fine-tuning approaches.
For the rest of our work, we select three best performing LM configurations with varying sizes -- (1) GPT-3 (with FEB template, and 6 randomly selected demonstrations), (2) T5-large (with infilling template, fine-tuned on the entire training set) and (3) T5-3B (with infilling template and 128-shot fine-tuning).
\paragraph{Task Performance.} For the three selected best performing LM configurations, we note (Tables \ref{tab:finetune_res}, \ref{tab:icl_res}) that task performance increases after the LM is forced to generate rationales. 
This is also consistent with prior findings \cite{cot, marasovic-etal-2022-shot}.

\subsubsection{Self-Rationalising Models Training Details}
In the experiments, we mainly used 3 models: T5-Large, T5-3B, and GPT-3 (model details and hyperparameters are shown in Table \ref{tab:model_details}). For T5-Large, we used the full train set for finetuning. For T5-3B, we trained in 2 settings: 48-shot and 128-shot. We used 3 seeds for generating shots for T5-3B. For GPT-3, we only used the OpenAI GPT-3 API \cite{brown2020language} to do inference.

\begin{table}[h!]
\centering
\scalebox{0.75}{
\begin{tabular}{cc}
\toprule
\textbf{Config}&\textbf{Assignment}\\
\midrule
models &
\makecell[c]{
\textbf{T5-3b}\\
Number of parameters: 3 billion
\\
\midrule
\textbf{T5-large}\\
Number of parameters: 770 million
\\
\midrule
\textbf{GPT3(davinci-instruct-beta)}\\
Number of parameters: 175 billion\\
\midrule
}\\
train batch size&4\\
eval batch size&4\\
seed&0\\
max epochs&25\\
learning rate&3e-5\\
learning scheduler&fixed\\
GPU&Quadro RTX 6000\\
Training time& 2 hours\\
\bottomrule
\end{tabular}
}
\caption{\textbf{Self-Rationalising Models Training Details}: Here we show the models we used and hyperparameters we used for T5-3B and T5-Large model training.}
\label{tab:model_details}
\end{table}

\subsection{Property Analysis}

For rationales generated by all three LMs, as well as gold rationales, we conduct human studies to evaluate whether the rationales satisfy the given properties.
For each instance, a property is marked on a binary scale (Yes / No), indicating the presence or absence of that property and evaluated by five annotators. 
Each category of properties is evaluated on a separate HIT, for which instructions have been modified so as to ensure that the annotators understand our definitions of the properties.
Given the complex nature of the human study, we make sure that the property annotations reach low to moderate agreement across all annotators (Table \ref{tab:property_annotation_agreement}).
% \brihi{Add agreement rate numbers}.

\begin{table*}[h]
\centering
\scalebox{0.75}{
\begin{tabular}{l|ccccccccc}
\toprule
\textbf{Rationale} & \multicolumn{1}{l}{\textbf{Grammaticality}} & \multicolumn{1}{l}{\textbf{Validity}} & \multicolumn{1}{l}{\textbf{Coherence}} & \multicolumn{1}{l}{\textbf{Conciseness}} & \multicolumn{1}{l}{\textbf{Leakage}} & \multicolumn{1}{l}{\textbf{Novelty}} & \multicolumn{1}{l}{\textbf{Association}} & \multicolumn{1}{l}{\textbf{Contrast}} & \multicolumn{1}{l}{\textbf{Average}} \\
\midrule
Gold & 0.11 & 0.18 & 0.19 & 0.10 & 0.24 & 0.21 & 0.12 & 0.24 & 0.17 \\
GPT-3 & 0.14 & 0.18 & 0.14 & 0.39 & 0.25 & 0.12 & 0.32 & 0.42 & 0.25 \\
T5-3B & 0.11 & 0.22 & 0.18 & 0.16 & 0.27 & 0.19 & 0.11 & 0.15 & 0.17 \\
T5-Large & 0.33 & 0.51 & 0.22 & 0.10 & 0.24 & 0.13 & 0.26 & 0.33 & 0.27 \\
\bottomrule
\end{tabular}}
\caption{\textbf{Annotation Agreements for Property Ratings}: Shown here are annotation agreements (Krippendorf's $\alpha$) for each property rating, along with aggregated agreements.}
\label{tab:property_annotation_agreement}
\end{table*}

\paragraph{Presence of properties in Gold and LM-generated Rationales}

\begin{figure*}[h!]
    \centering
    \includegraphics[width=\linewidth]{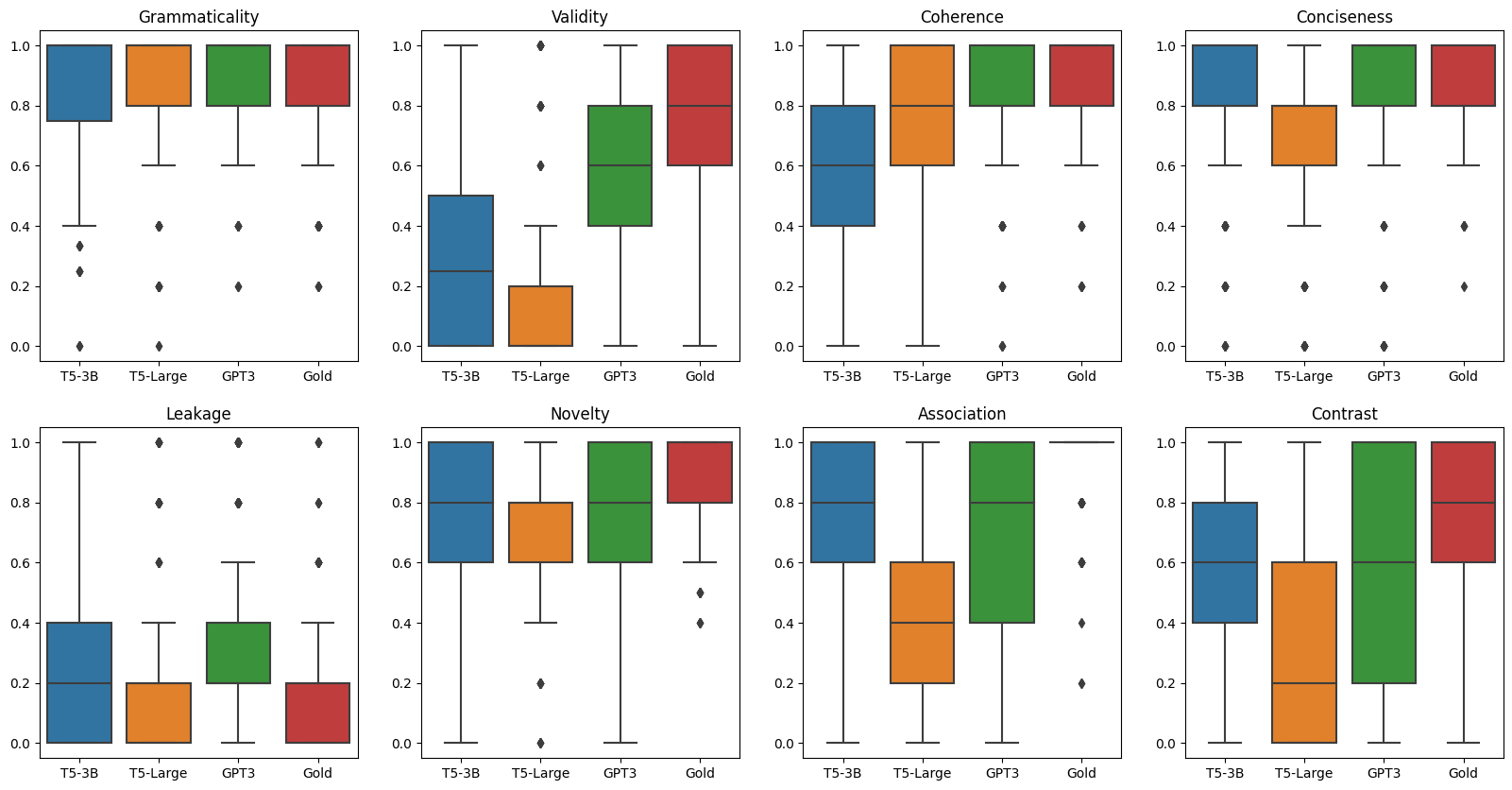}
    \caption{\textbf{Distribution of Property Annotations for Different Rationales:} Distribution is generated by aggregating scores of five annotators of each instance. A higher value implies more presence of the property in the rationale generated by the particular LM.}
    \label{fig:properties_distribution}
\end{figure*}

We first study the presence of these properties in rationales, without considering the utility of these rationales.
Figure \ref{fig:properties_distribution} plots the distribution of these properties, split by the models that generate these rationales, along with Gold rationales.
The distributions are obtained by taking the mean of ratings from five annotators for a given instance, where a higher value indicates a more frequent presence of that particular property in the set of rationales.
We observe that Gold rationales, in comparison to other model-generated rationales, have lower scores for leakage and higher scores for other properties.
In fact, Gold rationales are always associated with the gold label, which serves as a sanity check, as they are designed to help answer the gold label.
% \sahana{move sanity check to the end of the sentence.}
While all types of rationales are mostly grammatically correct
% \sahana{(change to) grammatically correct (also point to corresponding scores in table)}
, T5-Large and T5-3B suffer at producing rationales that are factually correct, and 
T5-Large rationales also tend to hallucinate and produce redundant sentences in rationales more often. 
% \sahana{In some places you say `the rationales' in the passive voice, and in other places you say `T5 produces rationale that is' in active voice. It might be good to standardize this across the whole paper.}
While GPT-3 rationales tend be generally better than T5-Large and T5-3B for surface-form and stylistic properties, they leak the predicted label more often than them. 
There is high variation for rationale-label association and contrasting features in rationales for all model-generated rationales, however on average, GPT-3 generated rationales are better on these metrics too. 
% \sahana{Point to some table/corroborating scores in all these sentences.}

\subsubsection{Property Correlations with Human Utility}
\label{sec:apx:prop_utility}

We use a Generalized Linear Mixed-Effects Model (GLMEM) (similar to \citet{lamm2020qed}) to model the correlation of different properties and their interactions with that of human utility. 
The formula used for modelling the GLMEM is as follows: 
$ \small \textsc{Response} = (\textsc{Grammaticality} + \textsc{Validity} + \textsc{Coherence} + \textsc{Conciseness} + \textsc{Leakage} + \textsc{Novelty} + \textsc{Association} + \textsc{Contrast})^2 + (1|\textsc{Question ID}) + (1|\textsc{Model ID}) + (1|\textsc{Human Prior})$

The response (dependent variable) is human accuracy after the human was shown the rationale. More formally,
$$ \small \textsc{Response} = \begin{cases} 
      1 & y_{hr} = \hat{y}\\
      0 & y_{hr} \neq \hat{y}\\
   \end{cases}
$$
All properties, along with their second-order interactions (implemented using the squared term above) are dependent variables.
Furthermore, we try to control for random effects whose variability might influence the response.
We control for randomness induced by a particular question, the model generating the rationales or whether the human had correctly answered the question before (Human Prior). More formally,
$$ \small \textsc{Human Prior} = \begin{cases} 
      1 & y_{h} = \hat{y}\\
      0 & y_{h} \neq \hat{y}\\
   \end{cases}
$$

\begin{table}[]
\centering
\scalebox{0.80}{
\begin{tabular}{ccc}
\toprule
\textbf{Property} & \textbf{Present} & \textbf{Absent} \\
\midrule
Grammaticality & -0.568 & -0.686 \\
Validity & -0.554 & -0.700 \\
Coherence & -0.665 & -0.589 \\
Conciseness & -0.540 & -0.714 \\
Leakage & -0.616 & -0.638 \\
Novelty & -0.712 & -0.542 \\
Association & -0.632 & -0.622 \\
Contrast & -0.613 & -0.641 \\
\bottomrule
\end{tabular}}
\caption{\textbf{Influence of individual properties in human utility:} Log odds of a rationale being useful, when a certain property is present or absent.}
\label{tab:property_individual}
\end{table}

Table \ref{tab:property_individual} shows the log odds of a rationale being useful when a certain property is present or absent, while averaging over other properties.
We note that all of the log odds are negative, which means that in isolation, the presence or absence of any property does not correlate well with rationales of high utility.

\begin{table}[h!]
\centering
\scalebox{0.80}{
\begin{tabular}{lc}
\toprule
\multicolumn{1}{c}{\textbf{Parameter}} & \textbf{Coefficient (SD)} \\
\midrule
\multicolumn{1}{l}{(Intercept)} & -0.724 (0.72) \\
+ grammaticality + leakage & 0.226 (0.55) \\
+ conciseness + novelty & 0.169 (0.32) \\
+ grammaticality + novelty & 0.149 (0.50) \\
+ coherence + novelty & 0.138 (0.23) \\
+ novelty + contrast & 0.136 (0.27) \\
+ conciseness + contrast & 0.119 (0.37) \\
+ validity + leakage & 0.118 (0.19) \\
+ association + contrast & 0.112 (0.54) \\
+ leakage + contrast & 0.098 (0.29) \\
+ coherence + association & 0.095 (0.27) \\
\bottomrule
\end{tabular}}
\caption{\textbf{Pairwise property interactions for rationale utility}: Given an intercept (when a rationale does not satisfy any property), the top ten pairs of properties that lead to an \textit{increase} in the log odds of a rationale being useful from the intercept is shown.}
\label{tab:property_interactions}
\end{table}

We then look at pairwise interactions. 
Table \ref{tab:property_interactions} shows the  top ten pairs which lead to an increase in utility log odds from the base level (Intercept), which is when a rationale does not satisfy any property.
A grammatically correct rationale that explicitly leaks the answer leads to the highest increase in log odds. 
This is also intuitive, as leakage is a direct signal to a human to select a given answer, without any reasoning from the human's behalf.

When all possible combinations of properties are considered, presence of all but coherence and association leads to a positive log odds for rationale utility: $0.139$.

\subsection{Quark training details}
For the Quark experiments, we used T5-Large as the self-rationalizing LM, and T5-Base for \reward. The hyperparameters used for running Quark \cite{lu2022quark} are shown in Table \ref{tab:quark_details}.

\begin{table}[h!]
    \centering
    \scalebox{0.80}{
    \begin{tabular}{ll}
    \hline 
    \textbf{Hyperparameter}   & \textbf{Value} \\ \hline
    Optimizer & Adam \\
    Adam epsilon & $1e$-$8$ \\
    Adam initial learning-rate & $1e$-$5$ \\
    Learning-rate scheduler & linear with warmup \\
    Warmup steps & 1000 \\
    Gradient clipping & $1.0$ \\
    Gradient accumulation & $2$ steps \\ \hline
    KL-divergence coef. & $0.05$ \\
    Entropy regularization coef. & $0.05$ \\ \hline
    \multirow{2}{*}{Sampling rate} & $2$ samples for \\
     & every train sample \\
     Frequency of exploration & every $500$ steps \\
     Sampling strategy & Top-p ($0.7$) sampling \\ 
     Temperature for sampling & $1.0$ \\ 
     Number of distinct reward-bins & $3$ ($1$, $0$ and $-1$) \\ \hline
    Train batch-size & $4$ \\
    Eval batch-size & $64$ \\ 
    Training time & $5$-$6$ hours \\ \hline
    \end{tabular}
    }
    \caption{Quark training details}
    \label{tab:quark_details}
\end{table}

\subsection{Examples}
In Table \ref{tab:generalization_demonstrations} we provide the demonstrations used to generate generalization questions using GPT-3. In Table \ref{tab:generalization_utility_examples}, we provide examples of useful, unsure and non-useful rationales with respect to human generalization. In Table \ref{tab:generalization_results} (corresponding to Figure \ref{fig:generalization_results}) we provide results for the difference in accuracies of human generalization, before and after a human annotator was shown the original question's rationale.

\subsection{MTurk Details}
\label{sec:apx:mturk}
\begin{figure}[h!]
    \centering
    \subfigure[\textbf{Instructions for human utility evaluation:} We first show annotators the description of the task and one example of HIT. We also included important notices to make sure annotators will use explanations.]{\frame{\includegraphics[width=0.9\linewidth]{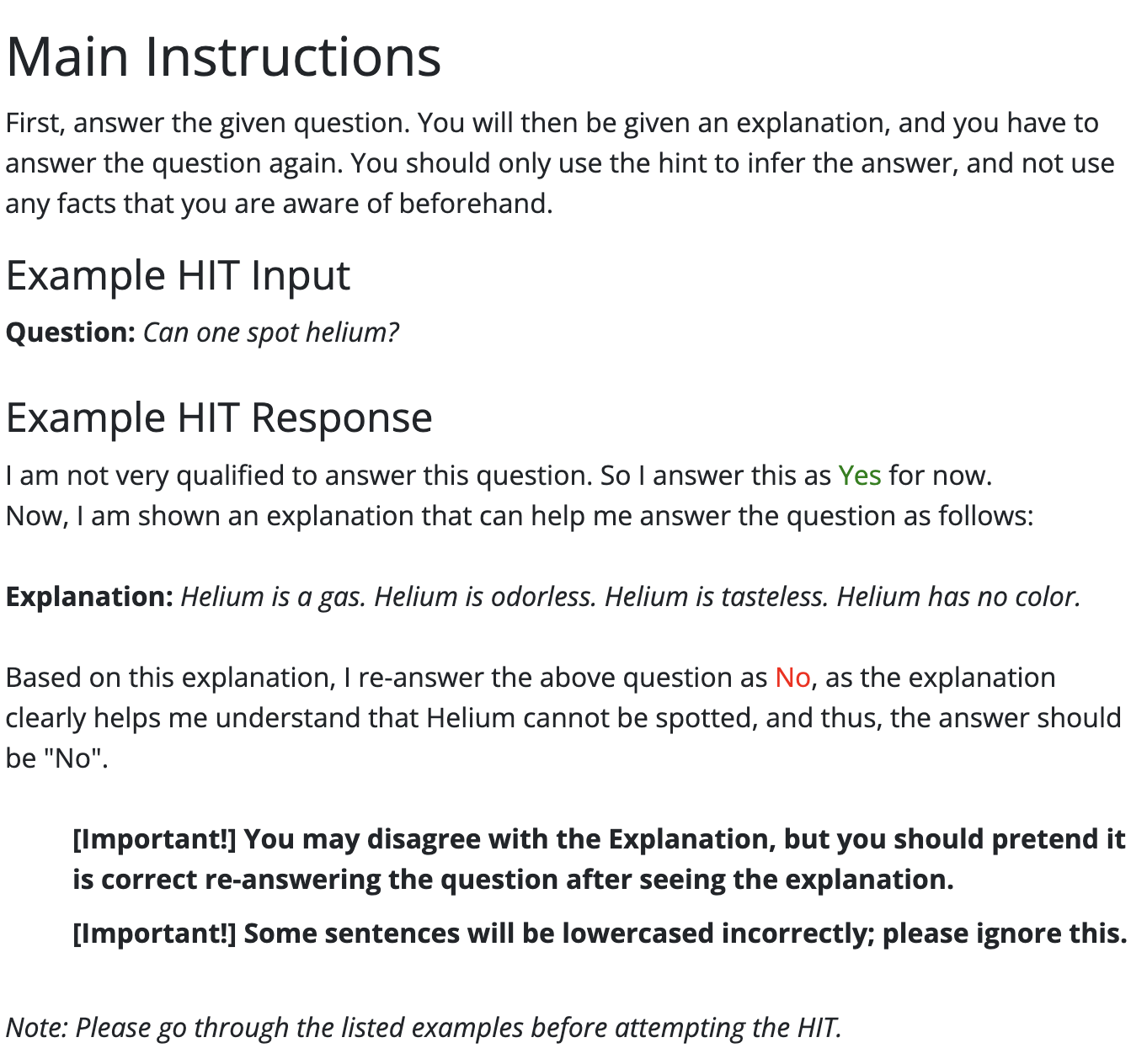}}}
    \subfigure[\textbf{An example for human utility evaluation:} We then show annotators 5 examples (we only show one of them in this figure). In the example, we will show them the procedure of annotations and how to response.]{\frame{\includegraphics[width=0.9\linewidth]{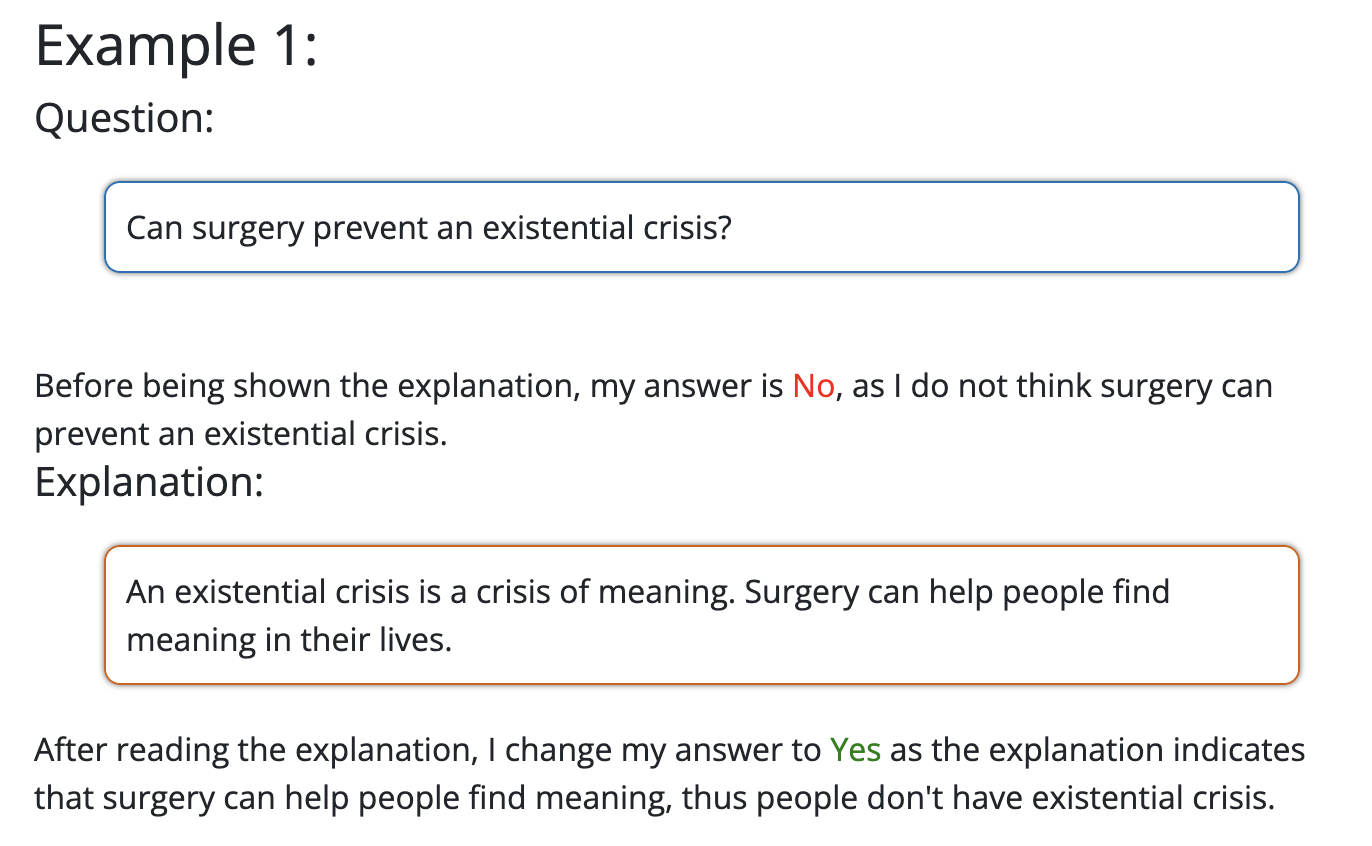}}}
    \subfigure[\textbf{Questionnaire for human utility evaluation:} Here is the template for evaluation. In the MTurk, the question and rationale will be replaced with real data. We will show the first question in the beginning. When annotators choose yes or no, the explanation and second question will appear.]{\frame{\includegraphics[width=0.9\linewidth]{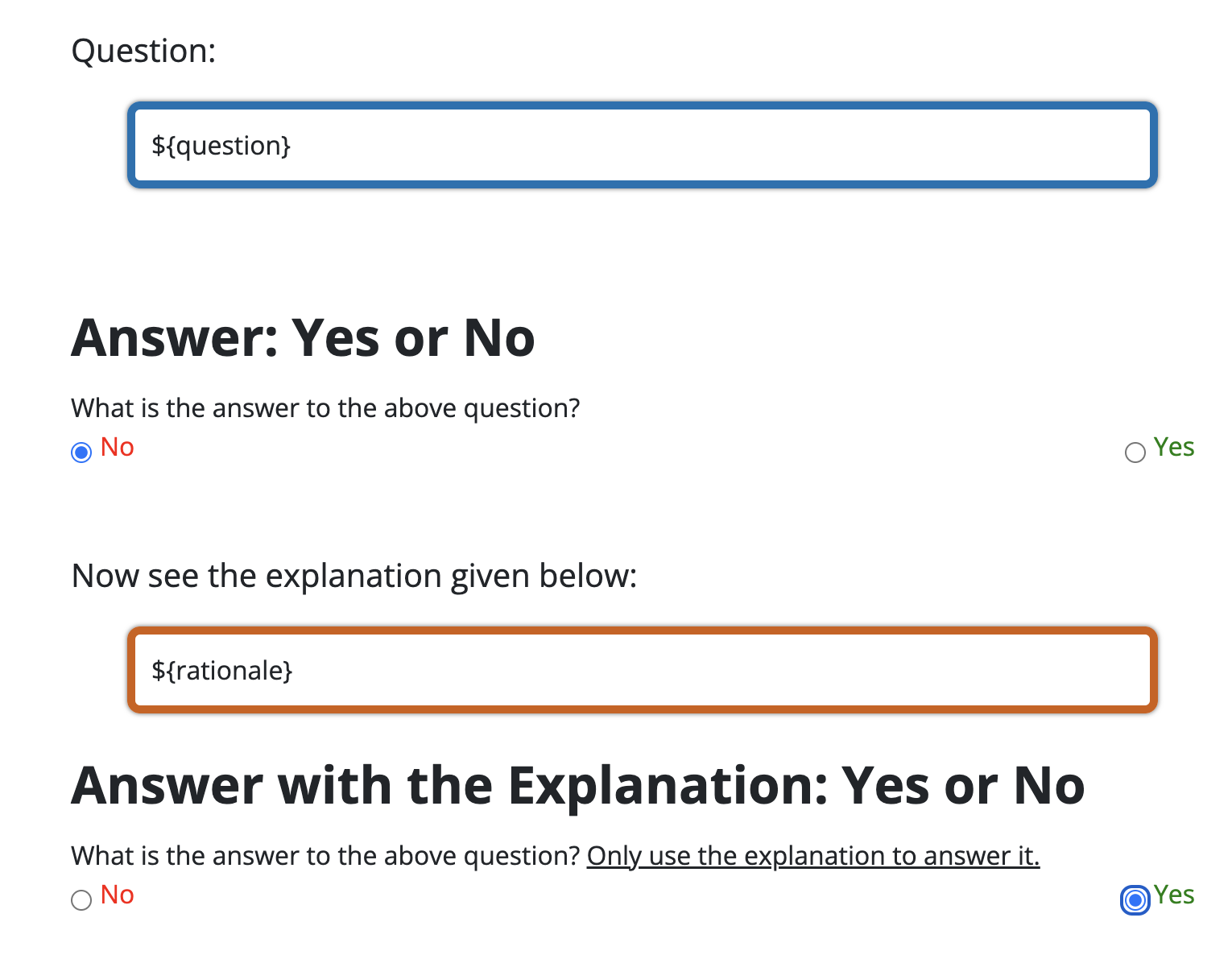}}}
    \caption{\textbf{The whole process for human utility evaluation}}
    \label{fig:fs_instruction_example_question}
\end{figure}

\begin{figure*}[h!]
    \centering
    \subfigure[\textbf{Instructions for property evaluation:} In this task, we split the property into 4 groups and conduct 4 rounds of annotations. (We show one of the groups - support).We rephrased 'label association' to 'support and 'contrast' to 'non-ambiguity' for easier understanding. In the introduction, we explain the properties and components of instances]{\frame{\includegraphics[width=0.45\linewidth]{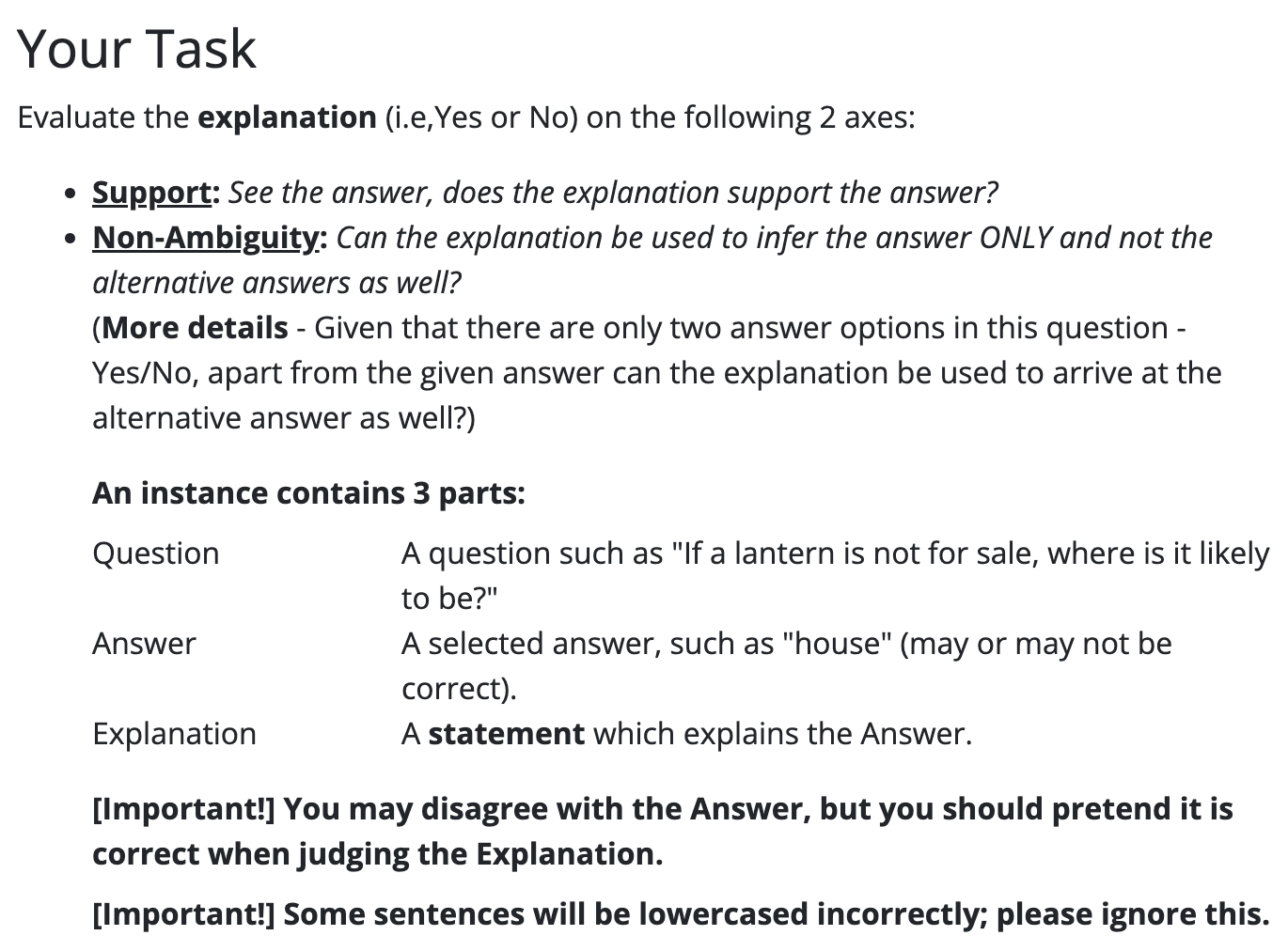}}} \hfill
    \subfigure[\textbf{Instructions for property evaluation:} In the instruction, we also include one HIT example. We explain the properties by showing negative examples.]{\frame{\includegraphics[width=0.45\linewidth]{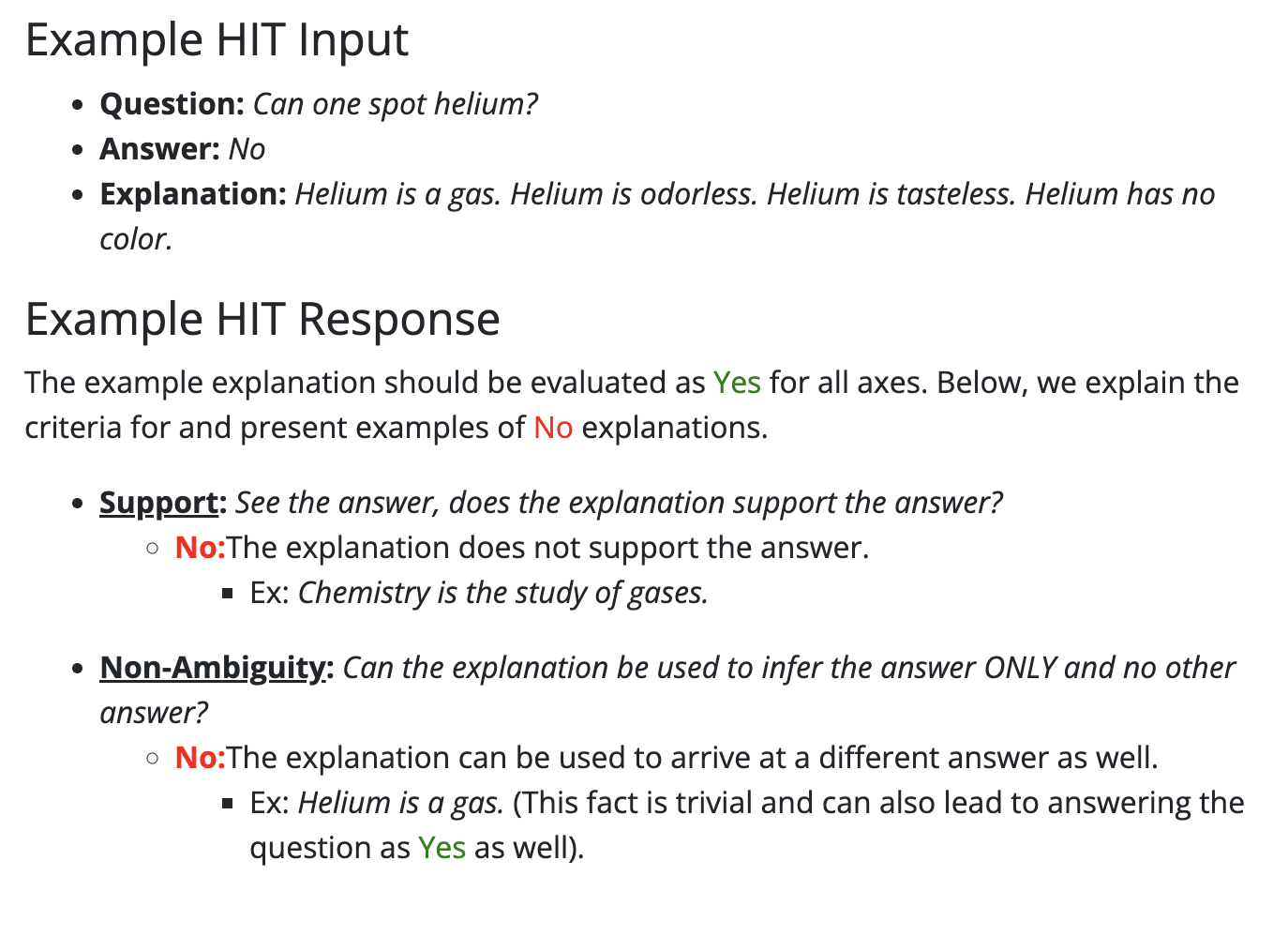}}}
    \subfigure[\textbf{An example for property evaluation:} We demonstrate 6 examples in the template and we show different combination of results in examples.]{\frame{\includegraphics[width=0.45\linewidth]{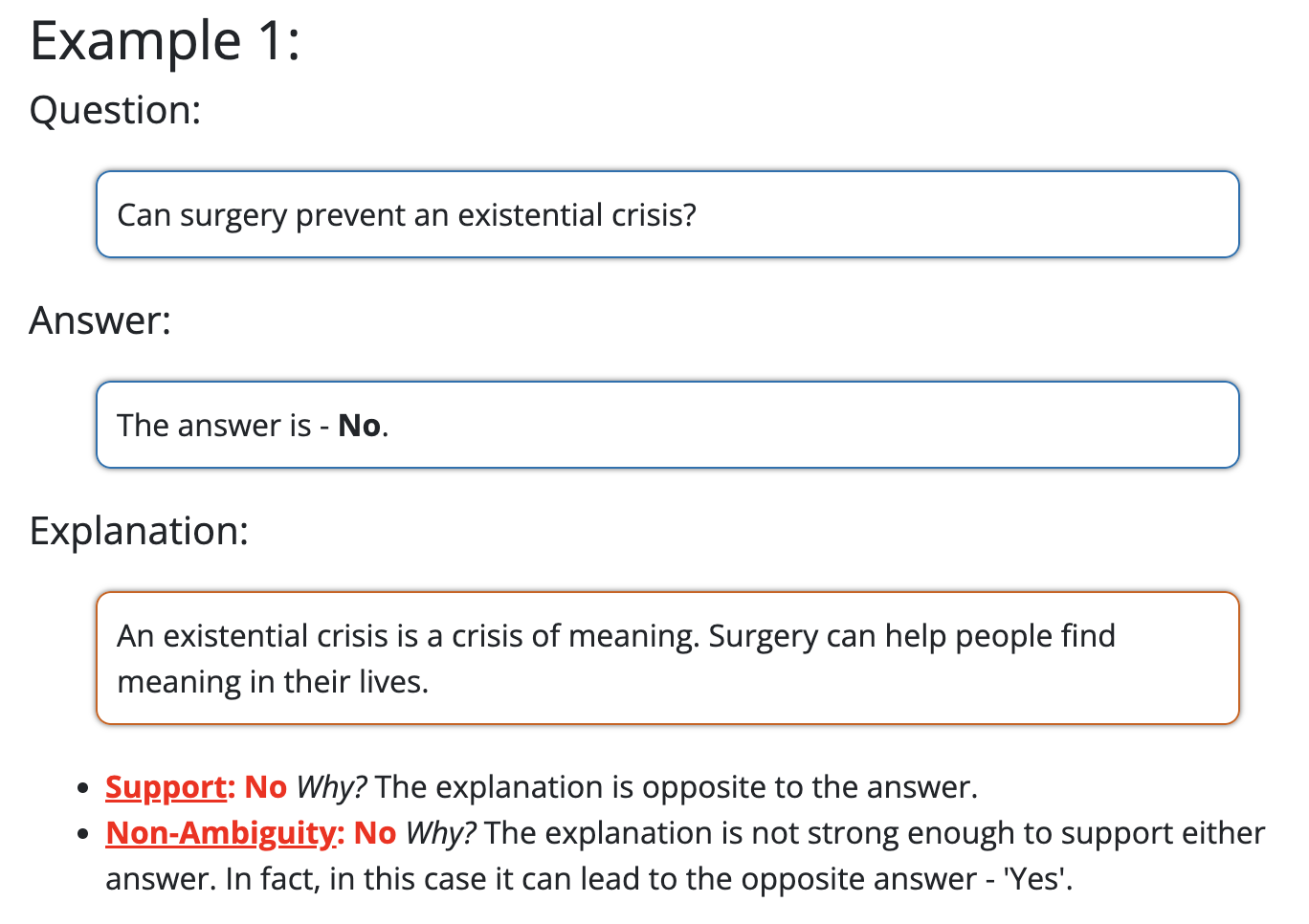}}} \hfill
    \subfigure[\textbf{Questionnaire for property evaluation:} Annotators will be shown a triplet of question, answer and explanation. Similar as the previous task, user need to answer the first question to get to the second one.]{\frame{\includegraphics[width=0.45\linewidth]{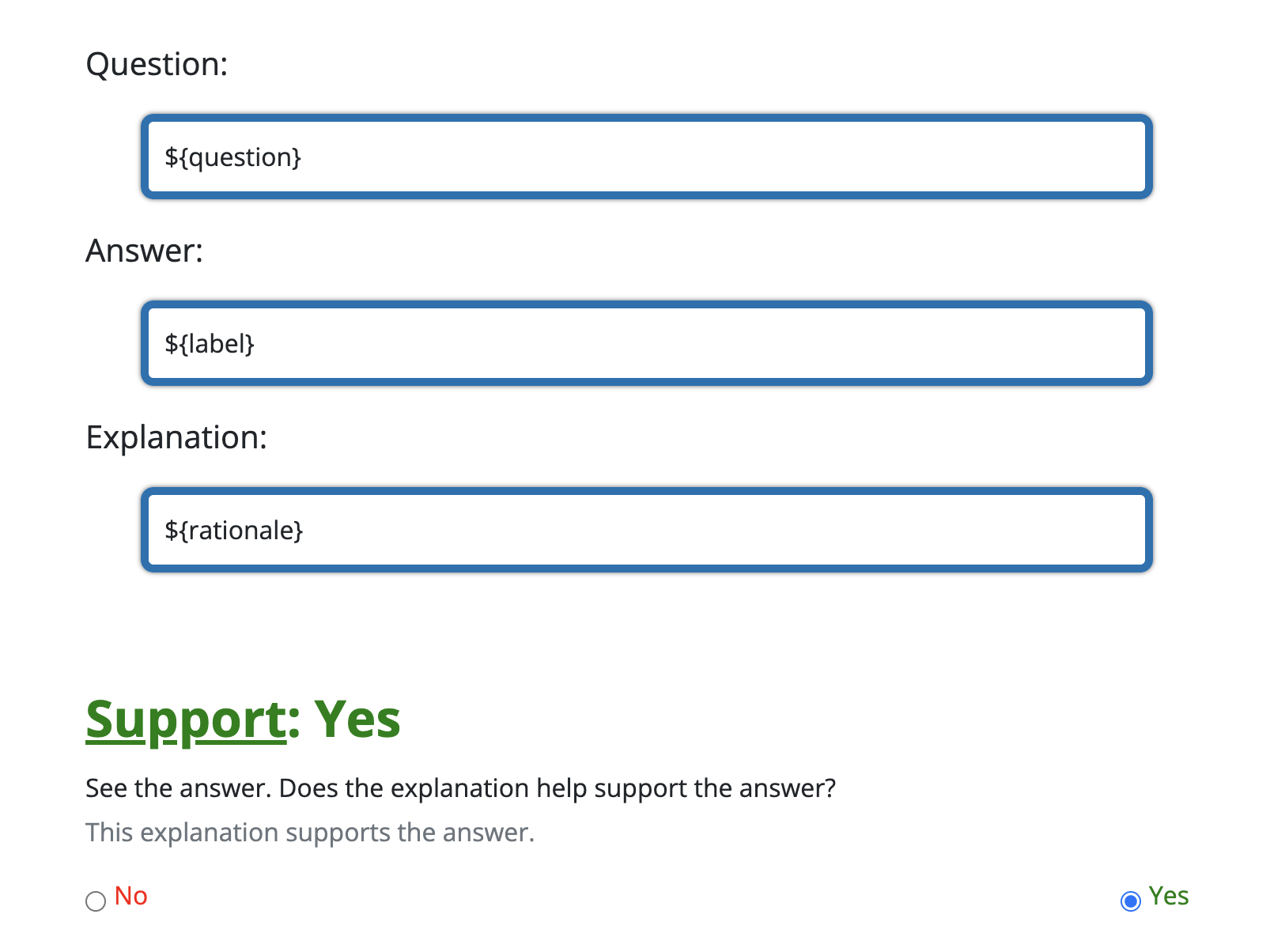}}}
    \caption{\textbf{One example of property evaluation questionnaire}: For other properties, they have the similar templates, with different instructions and examples.}
    \label{fig:property_instruction_figs}
\end{figure*}

\begin{figure*}[h!]
\centering
\subfigure[\textbf{Instruction for validation of generalization questions (similar reasoning):} We asked the annotators to validate if the related question is a similar reasoning question.]{
\frame{\includegraphics[width=0.45\linewidth]{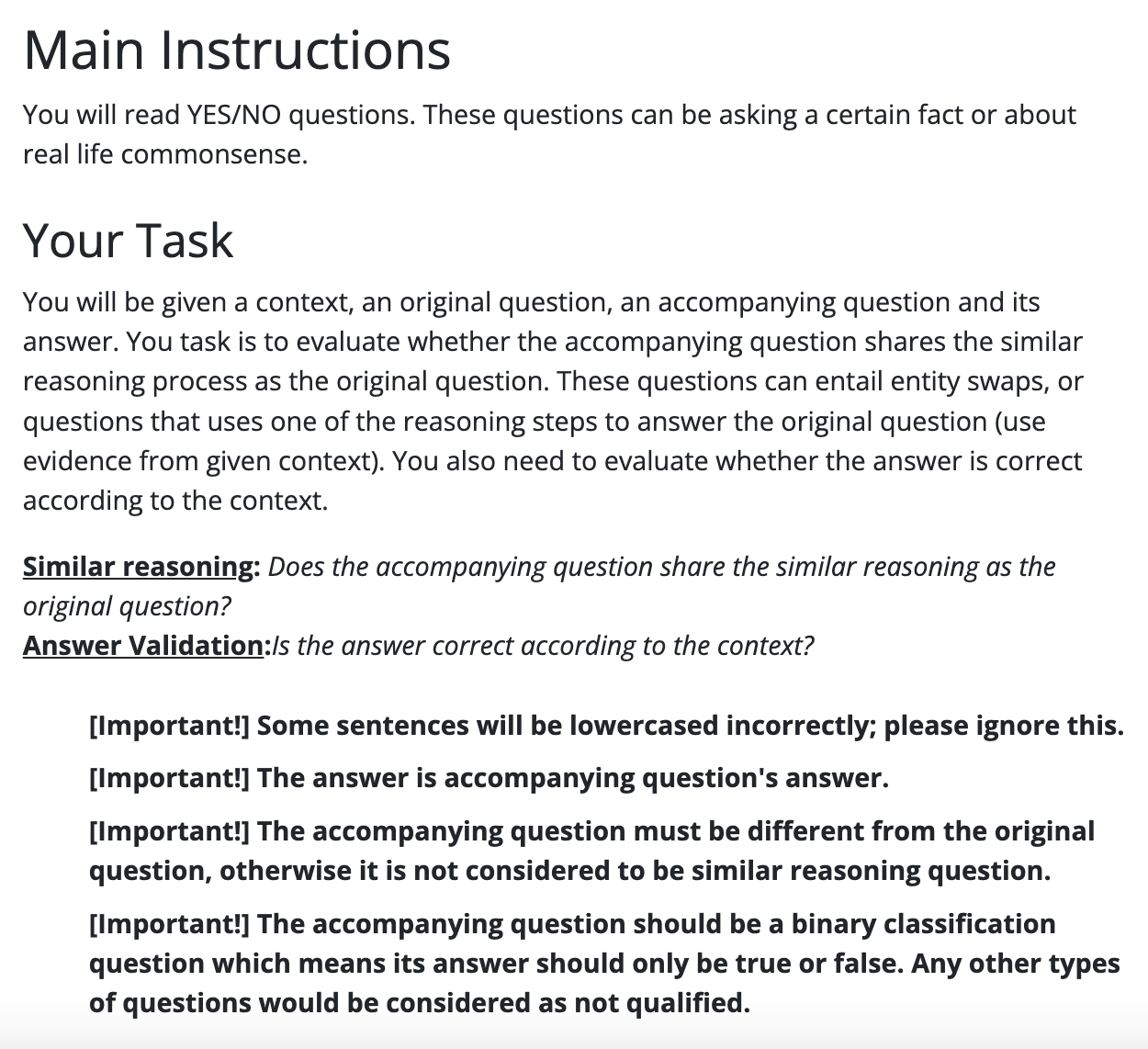}}\label{fig:filter_instruction}}\hfill
\subfigure[\textbf{Example for validation of generalization questions (similar reasoning):} We selected 3 examples in the template to clarify the definition of similar reasoning.]{\frame{\includegraphics[width=0.45\linewidth]{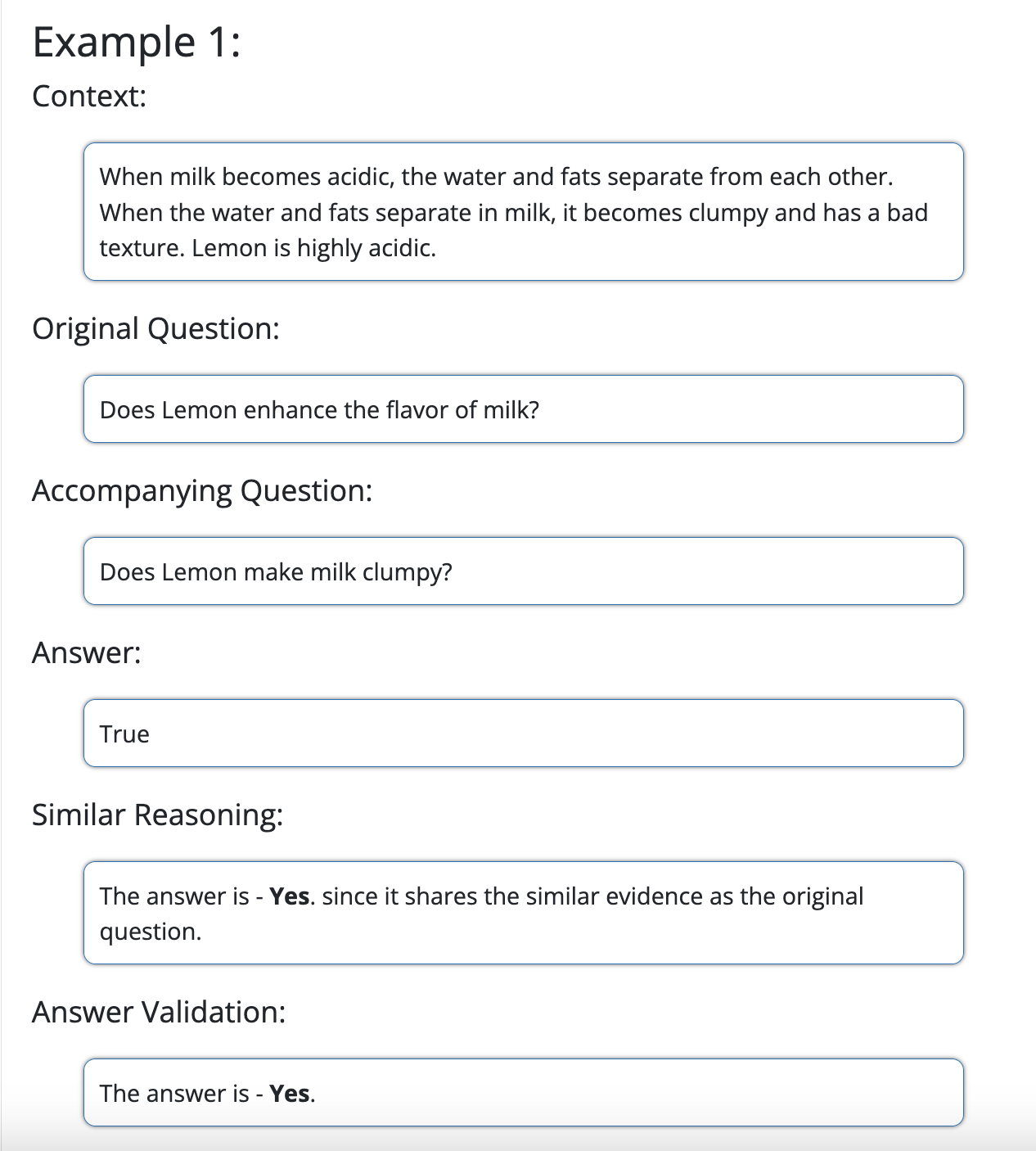}}}
   \subfigure[\textbf{Questionnaire for validation of generalization questions (similar reasoning):} In the questionnaire, annotators need to validate whether the related question is a similar reasoning question then validate the answer of the related question.]{\frame{\includegraphics[width=0.45\linewidth]{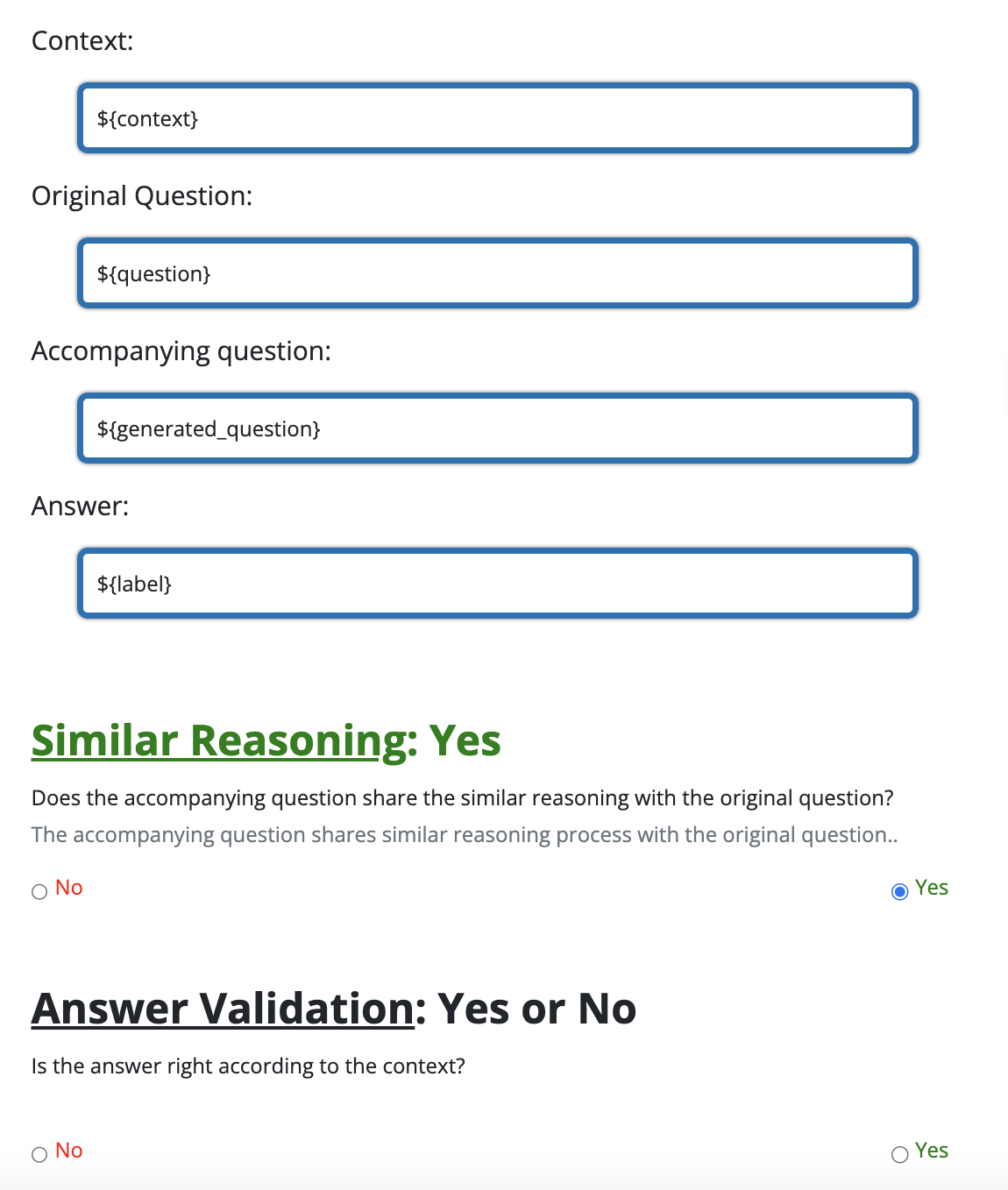}}}
    
% \hfill
%     \subfigure[\textbf{Instruction for generalization question:} In section 5, generalization questions are divided into 3 types, but in MTurk, we hide this information from annotators. Instruction will help annotators to understand the process and what is follow-up question.]{\frame{\includegraphics[width=0.45\linewidth]{figures/gen_instruction.png}}\label{fig:gen_instruction}}
    \caption{\textbf{Validation of generalization question}: Rephrase and counterfactual have the similar setup, except for the answer validation. We assume that rephrase questions should have the same answer of original ones and the counterfactual questions should have the opposite answer.}
    \label{fig:filter_gen}
\end{figure*}

\begin{figure*}[h!]
    \centering
        \subfigure[\textbf{Instruction for generalization question:} In section 5, generalization questions are divided into 3 types, but in MTurk, we hide this information from annotators. Instruction will help annotators to understand the process and what is follow-up question.]{\frame{\includegraphics[width=0.45\linewidth]{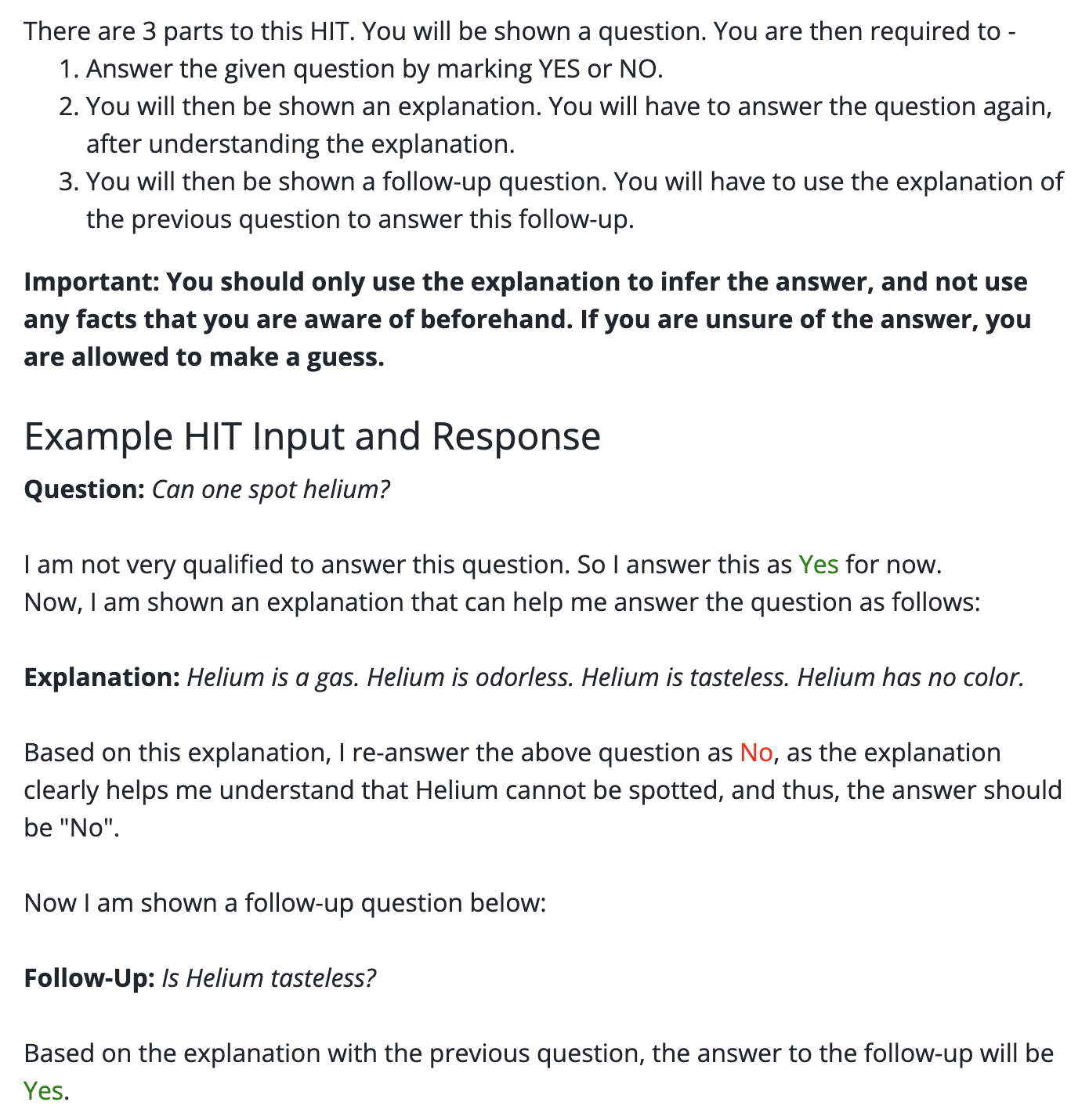}}\label{fig:gen_instruction}}\hfill
        \subfigure[\textbf{An example for generalization question:} We demonstration 5 examples in the template. We show how our thinking process change before and after given explanation and how explanation help to answer the follow-up question.]{\frame{\includegraphics[width=0.45\linewidth]{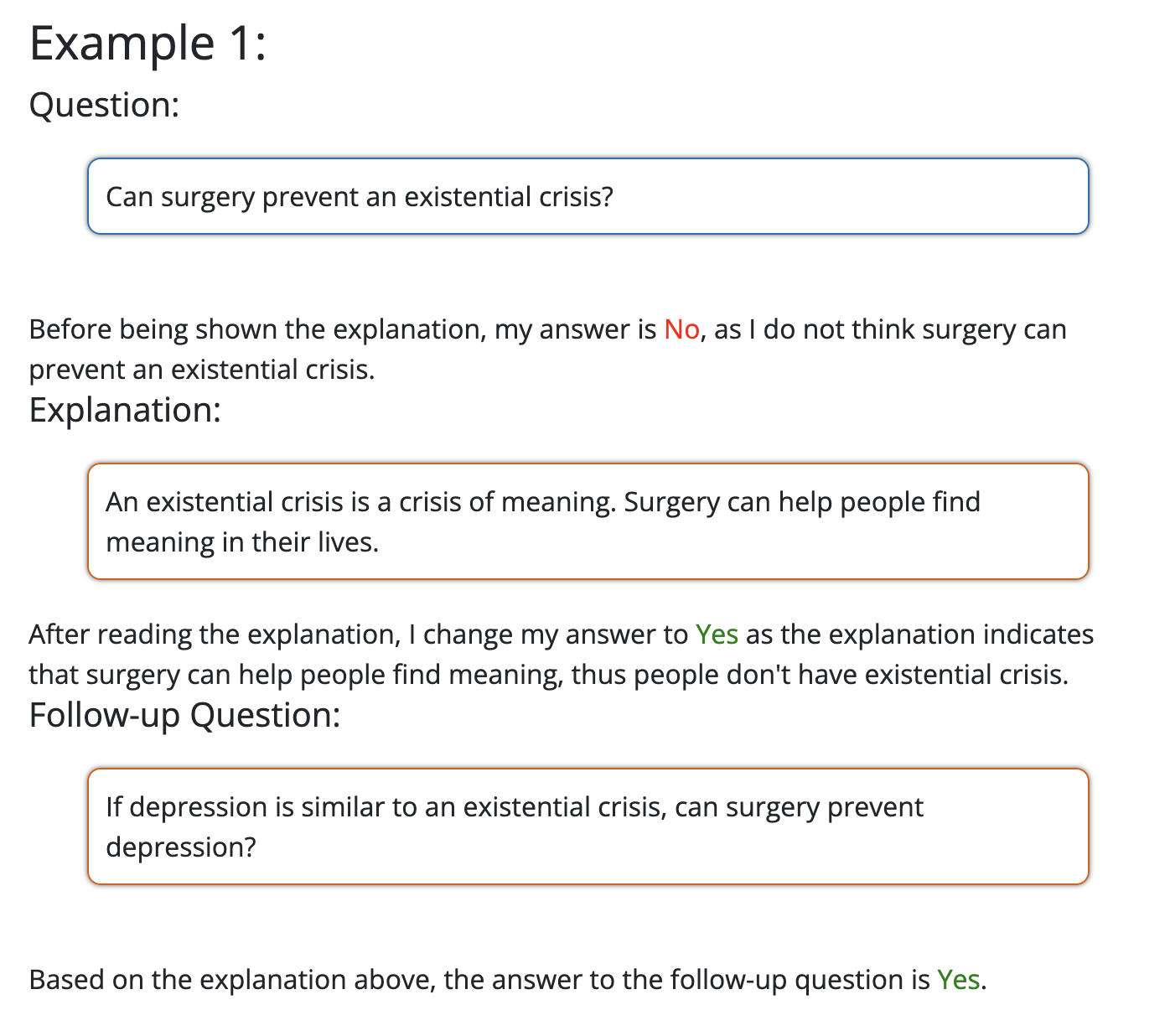}}}
        \subfigure[\textbf{Questionnaire for generalization question:} In the questionaire, annotators will repeat the steps in human utility evaluations. We repeat it because we cannot make sure annotators took human utility evaluations and annotators took generalization question evaluations will be same group of people. After this, we show them follow-up question and ask them to use the explanation to answer the question.]{\frame{\includegraphics[width=0.45\linewidth]{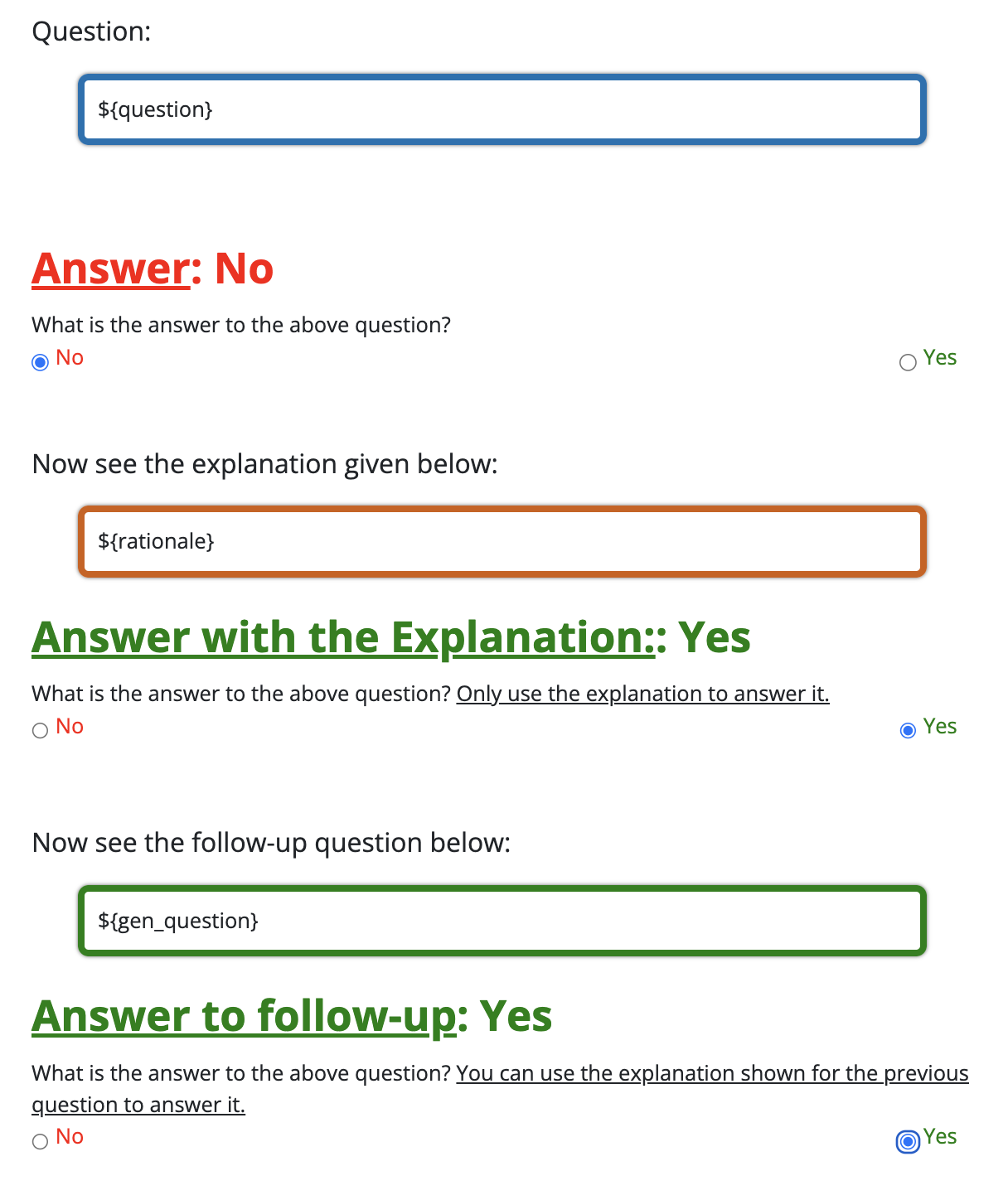}}} 
    % \subfigure[\textbf{Questionnaire for validation of generalization questions (similar reasoning):} In the questionnaire, annotators need to validate whether the related question is a similar reasoning question then validate the answer of the related question.]{\frame{\includegraphics[width=0.45\linewidth]{figures/filtering_questionaire.png}}} \hfill
    % \subfigure[\textbf{Example for validation of generalization questions (similar reasoning):} We selected 3 examples in the template to clarify the definition of similar reasoning.]{\frame{\includegraphics[width=0.45\linewidth]{figures/filtering_example.png}}}
    \caption{\textbf{Measuring rationale utility by answering generalization questions}}
    \label{fig:generlization_example_question}
\end{figure*}

In this section, we describe the MTurk experiment setup. The details of MTurk experiments including how many Turkers took the evaluation, and average time used to finish evaluations are shown in Table \ref{tab:appendix_turk}. 
Each MTurk annotator is paid above minimum wage. 
Figure \ref{fig:fs_instruction_example_question} demonstrates the setup for human utility evaluation.
Figure \ref{fig:property_instruction_figs} demonstrates the setup for property evaluation. 
Figures \ref{fig:filter_gen} demonstrates the setup for validating generalization questions. 
Figure \ref{fig:generlization_example_question} demonstrate the setup for utility evaluation towards generalization questions.\\
Since the dataset we used is carefully annotated by human, we can assure there is no toxic content and our experiment setup was submitted to IRB for ethical review. 
% We assure that each Turker can get fair pay comparing to market price.
We limited our Turkers to English speaking nations - United States, Canada, Australia, New Zealand and United Kingdom. \\
To ensure the quality of evaluation, we did a round of qualification task before each task which include a small set of evaluations. 
Turkers need to finish the qualification task first and get results of it, then we will show them the whole task.\\

\subsubsection{Worker Selection and Quality Control}

Here, we describe details about how workers are selected and how annotations are ensured to be clean. 
First, we employ multiple rounds of trials before deploying the actual task so as to get feedback from annotators whether they understand the task correctly.
This includes in-house tests, tested via Amazon Turk Sandbox \footnote{\url{https://requester.mturk.com/developer/sandbox}} and small batches tested on Turk.
Second, we create a set of medium to hard qualification tasks for each task that the annotators have to work on. 
These tasks are hand curated that cater certain parts of the instruction -- whether the annotators are reading the rationale correctly, or whether they are able to make appropriate connectections between the rationale and the question.
This weeds out a lot of annotators who do not understand the task or are cheating.
We also weed out workers who are too `fast' (completing the task in less than $5$ seconds, which is indicative of potential slacking in the task).
Third, we constantly monitor task responses and feedback provided to annotators about their task. 
We also collect feedback from them which we adapt in new versions of the task. 

\subsubsection{Turking for Generalization Questions}
\label{sec:turk_gen_question}

Each generalization question is validated by $3$ annotators each.
The validation process includes: checking if the generated question can be answered by the gold rationale, answering the generated question, and checking if the generated question follows the instructions for a given type (being a rephrase, counterfactual or a similar reasoning question). 
The annotation agreement observed here is high (Krippendorf's $\alpha=0.68$).

\subsubsection{Annotation Agreements}
We observe that StrategyQA instances are difficult to annotate by humans, as many of them are fact-based, which the human might or might not know beforehand.
Therefore, human agreement \textit{before} the rationale is shown is low (Krippendorf's $\alpha=0.18$). 
However, \textit{after} being shown the rationale, the agreement increases, as shown in Table \ref{tab:human_agreement}.
% with Krippendorf's $\alpha$ being $0.47$, $0.30$ and $0.24$ for GPT-3, T5-3B and T5-Large LMs respectively.
Examples of rationales annotated into each of the three human utility categories (useful, not useful, unsure) is shown in Table \ref{tab:examples_utility}.

\begin{table}[h!]
\centering
\scalebox{0.65}{
\begin{tabular}{ccc}
\toprule
\textbf{Tasks}&\textbf{Number of Turkers} &\textbf{Average Time(s)} \\
\midrule
Human Utility Evaluation & 80 & 37.41\\
Property Evaluation & 137 & 36.50\\
Generalization Question & 25 & 35.93\\
\bottomrule

\end{tabular}

}
\caption{\textbf{Details of MTurk:} Shown here are number of unique Turkers (annotators) and average time of solving one HIT for each task}
\label{tab:appendix_turk}
\end{table}

\begin{table}[t!]
\centering
\scalebox{0.85}{
\begin{tabular}{c|ccc}
\toprule
Model & GPT-3 & T5-3B & T5-Large\\
\midrule
Krippendorf's $\alpha$ & 0.47 & 0.30 & 0.24 \\
\bottomrule
\end{tabular}
}
\caption{\textbf{Annotators agreement}:Shown here is the annotators agreement. davinci-instruct-beta (GPT-3) has the best agreement even though its task performance is low. Contrastly, T5-Large has highest task performance but a low agreement.}
\label{tab:human_agreement}
\end{table}

% \begin{figure*}[h!]
%     \centering

%     \caption{Caption}
%     \label{fig:gen_example_question}
% \end{figure*}

\begin{table*}[h]
\centering
\scalebox{0.65}{
\begin{tabular}{ll}
\toprule
\textbf{Utility}&\textbf{Examples}\\
\midrule
\textbf{Useful}&
\makecell[l]{
\textbf{Original Question}:Did Evander Holyfield compete in an Olympics hosted in the western hemisphere?
\\
\textbf{Rationale}:Evander Holyfield competed in the 1984 Olympics in Los Angeles, California.\\
\textbf{Generalization Question}:Did Evander Holyfield compete in an Olympics hosted in the United States?\\
\midrule
\textbf{Original Question}:Is Nine Inch Nails's lead singer associated with David Lynch?
\\
\textbf{Rationale}:Nine Inch Nails's lead singer is Trent Reznor. Trent Reznor is a producer for David Lynch's film, "Split" (1985). \\David Lynch produced Split (1985).
\\
\textbf{Generalization Question}:Is Trent Reznor associated with David Lynch?
}\\
\midrule
\textbf{Unsure}&
\makecell[l]{
\textbf{Original Question}:Is a beard is moss that grows on a human?
\\
\textbf{Rationale}:A beard is hair that grows on a human. Moss is a type of plant.\\
\textbf{Generalization Question}:Is a beard a type of plant?\\
\midrule
\textbf{Original Question}:Does the Red Sea have biblical significance?
\\
\textbf{Rationale}:The Red Sea is a body of water in the middle of the desert. The biblical story of Moses crossing \\ the Red Sea is found in Exodus 14:26-27. 
\\
\textbf{Generalization Question}:Is the Red Sea a biblical sea?
}\\
\midrule
\textbf{Not Useful}&
\makecell[l]{
\textbf{Original Question}:Has a baby ever had a moustache?
\\
\textbf{Rationale}:Babies are born without facial hair.\\
\textbf{Generalization Question}:Has a baby ever had lanugo?\\
\midrule
\textbf{Original Question}:Can Michael Jordan become a professional cook in America?
\\
\textbf{Rationale}:Michael Jordan was born in 1964 The United States of America was founded in 1776.\\
\textbf{Generalization Question}:Can Michael Jordan become a culinary apprentice?
}\\
\bottomrule
\end{tabular}

}
\caption{\textbf{Examples of rationales for Section 3}: For useful and unsure rationales, we selected those that support humans to answer the generalization questions correctly; and for not useful rationales, we selected examples where human failed to give the right answer.}
\label{tab:generalization_utility_examples}
\end{table*}
\begin{table*}[h!]
\centering
\scalebox{0.85}{
\begin{tabular}{ccccc}
\toprule
 &  & \multicolumn{3}{c}{\textbf{Generalization Accuracy}} \\
 \cmidrule(lr){3-5}
\textbf{Type of Generalization Questions} & \textbf{Model} & \textbf{Useful} & \textbf{Non-useful} & \textbf{Unsure} \\
\midrule
\multirow{5}{*}{Rephrase} & Gold & \cellcolor{lred} 94.68 & 34.24 & 94.35 \\
 & GPT-3 & 69.38 & 18.95 & 87.90 \\
 & T5-3B & 73.58 & 27.82 & 93.90 \\
 & T5-Large & \cellcolor{lblue} 74.11 & 25.60 & 90.00 \\
\cmidrule(lr){2-5}
 & Combined (Models) & 72.31 & 24.31 & 90.52 \\
\midrule
\multirow{5}{*}{Counterfactuals} & Gold & \cellcolor{lred} 79.50 & 57.34 & 71.83 \\
 & GPT-3 & \cellcolor{lblue} 75.00 & 43.47 & 62.11 \\
 & T5-3B & 57.57 & 39.72 & 50.22 \\
 & T5-Large & 70.66 & 35.06 & 52.45 \\
 \cmidrule(lr){2-5}
 &  Combined (Models) & 68.20 & 39.26 & 55.03 \\
\midrule
\multirow{5}{*}{Similar Reasoning} & Gold & \cellcolor{lred} 74.38 & 54.34 & 90.27 \\
 & GPT-3 & \cellcolor{lblue} 51.63 & 36.61 & 74.68 \\
 & T5-3B & 41.93 & 36.77 & 70.22 \\
 & T5-Large & 43.61 & 42.11 & 70.00 \\
 \cmidrule(lr){2-5}
 & Combined (Models) & 45.69 & 38.54 & 71.77 \\
\bottomrule
\end{tabular}
}
\caption{\textbf{Generalization Results} - Numbers corresponding to Figure \ref{fig:generalization_results}.}
\label{tab:generalization_results}
\end{table*}

\begin{table*}[h]
\centering
\scalebox{0.62}{
\begin{tabular}{lll}
\toprule
    \textbf{Category,Instruction} & \textbf{Demonstrations}  \\\midrule
    \makecell[l]{\colorbox{lpurple}{
    Rephrase}:\\
    Rephrase the question 
    \\and answer it.
    }
    &
    \makecell[l]{
    \textbf{question}:Are more people today related to Genghis Khan than Julius Caesar?\\\textbf{rephrase}:Do more people today have connection with Genghis Khan than Julius Caesar?\\\textbf{answer}:True.\\
    \midrule
   \textbf{question}:Would a dog respond to bell before Grey seal?\\\textbf{rephrase}: Would Grey seal respond to bell later than a dog?\\\textbf{answer}:True.\\
   \midrule
    \textbf{question}:Is a Boeing 737 cost covered by Wonder Woman (2017 film) box office receipts?\\\textbf{rephrase}:Does Wonder Woman box office receipts cover a Boeing 737 cost?\\\textbf{answer}:True.\\
    \midrule
    \textbf{question}:Is the language used in Saint Vincent and the Grenadines rooted in English?\\\textbf{rephrase}: Does the language used in Saint Vincent and the Grenadines originate from English?
    \\\textbf{answer}:True.\\
    \midrule
    \textbf{question}:Are Christmas trees dissimilar to deciduous trees?\\\textbf{rephrase}:Are Christmas trees different from deciduous trees?\\\textbf{answer}:True.\\
    \midrule
   \textbf{question}:Does Dragon Ball shows and movies fall short of Friday 13th number of projects?\\\textbf{rephrase}:Does Dragon Ball make less shows and movies than Friday 13th?
   \\\textbf{answer}:True
    } \\
    \midrule 
    \makecell[l]{\colorbox{lorange}{Counterfactual}:\\Given the context and question,\\ generate a question \\that negates the question.}&
    \makecell[l]{\textbf{context}:A plum tree is a deciduous tree that bears fruit. Deciduous trees shed their leaves in the\\ autumn. Autumn happens from September until the end of Deember.\\\textbf{question}:Is November a bad time for a photographer to take pictures of a plum tree in bloom?\\\textbf{generate}:Is a plum tree in bloom in the autumn?.\\
    \midrule
    \textbf{context}:The animals that Yetis are said to look similar to are able to use their hands or toes to \\grasp items The ability to grasp with hands or other limbs is to be prehensile. \\\textbf{question}:Would a Yeti be likely to have prehensile limbs?\\\textbf{generate}:Is a Yeti able to grasp items with its hands or toes?\\
    \midrule
    \textbf{context}:Keelhauling was a severe punishment whereby the condemned man was dragged beneath\\ the ship\u2019s keel on a rope. Keelhauling is considered a form of torture. \\Torture is considered cruel. The Eighth Amendment forbids the use of cruel and unusual punishment \\
    \textbf{question}:Would keelhauling be a fair punishment under the Eighth Amendment?\\\textbf{generate}:Would keelhauling be considered cruel?
    \\
    \midrule
    \textbf{context}:Khanbaliq was the winter capital of the Mongol Empire. Khanbaliq was located at the \\center of what is now modern day Beijing, China. Moon Jae-In was born in Geoje, South Korea.\\\textbf{question}:Was Moon Jae-in born outside of Khanbaliq?\\\textbf{generate}:Was Moon Jae-in born in Beijing?\\
    \midrule
    \textbf{context}:Amazonas is mostly tropical jungle. Tropical jungles contain dangerous creatures. Dangerous \\creatures put people's lives at risk.\\\textbf{question}:Does walking across Amazonas put a person's life at risk?\\\textbf{generate}:Is Amazonas a safe place?\\
    \midrule
    \textbf{context}:The Los Angeles Memorial Sports Arena had a capacity of 16,740 people. Coachella has had \\attendance numbers in excess of 99.000 people. Coachella relies on an outdoor set up to accommodate\\ the massive crowds.\\\textbf{question}:Was Los Angeles Memorial Sports Arena hypothetically inadequate for hosting Coachella?\\\textbf{generate}:Would Los Angeles Memorial Sports Arena be too big for Coachella?}\\
    \midrule
     \makecell[l]{\colorbox{lgreen}{Similar reasoning}:\\Given a context, generate \\
    a similar question to the \\
    given question and answer it}&
     \makecell[l]{\textbf{context}:A plum tree is a deciduous tree that bears fruit. Deciduous trees shed their leaves in the autumn.\\ Autumn happens from September until the end of Deember.\\\textbf{question}:Is November a bad time for a photographer to take pictures of a plum tree in bloom?\\\textbf{generate}:Will the leaves a plum tree fall in the autumn?\textbf{answer}:True\\
    \midrule
    \textbf{context}:The Alamo is located in San Antonio. The Alamo was the site of a major battle during the \\Texan Revolution against Mexico in 1836.
    \\\textbf{question}:Was San Antonio the site of a major battle in the 19th century?
    \\\textbf{generate}:Was the Alamo the site of a major battle in the 19th century?\textbf{answer}:True\\
     \midrule
    \textbf{context}:Filicide is the act of killing a son or a daughter. Marvin Gay Sr. committed filicide in 1984 \\when he shot his son, singer Marvin Gaye. Isaac's father Abraham, was commanded by God to\\ sacrifice his son Isaac, but was spared by an angel.\\
    \textbf{question}:Did Isaac's father almost commit similar crime as Marvin Gay Sr?\\\textbf{generate}:Did Isaac's father almost commit filicide?\textbf{answer}:True
    \\
    \midrule
    \textbf{context}:The animals that Yetis are said to look similar to are able to use their hands or toes to grasp items.\\ The ability to grasp with hands or other limbs is to be prehensile.\\\textbf{question}:Would a Yeti be likely to have prehensile limbs?\\\textbf{generate}:Will a Yeti fail to grasp items with its hands or toes?\textbf{answer}:True\\
    \midrule
    \textbf{context}:Land of Israel was controlled by the Ottoman Empire in 16th century. The religion of Ottoman\\ Empire was Sunni Islam. \\\textbf{question}:Was Land of Israel in possession of an Islamic empire in 16th century?\\\textbf{generate}:Was the Ottoman Empire Islamic once?\textbf{answer}:True\\
    \midrule
    \textbf{context}:Wedding rings are typically made of precious shiny stones such as diamonds. Silicon is a solid\\ rock like element at room temperature that has a natural lustre. Bromine is a liquid at room\\ temperature that is toxic to the touch.\\\textbf{question}:Will silicon wedding rings outsell bromine wedding rings?\\\textbf{generate}:Are silicon wedding rings shiny?\textbf{answer}:True\\}\\

\bottomrule
\end{tabular}}
\caption{\textbf{Demonstrations for generating generalization questions}: For each category, we used 6 fixed demonstrations. We used different questions for each category. }
\label{tab:generalization_demonstrations}
\end{table*}

\end{document}